\documentclass[runningheads]{llncs}
\usepackage[final,year=2024,ID=1255]{eccv}
\usepackage{eccvabbrv}
\usepackage{graphicx}
\usepackage{booktabs}
\usepackage[accsupp]{axessibility}
\usepackage[pagebackref,breaklinks,colorlinks]{hyperref}
\usepackage{orcidlink}
\usepackage{multirow}
\usepackage{amsmath}
\usepackage{amssymb}
\usepackage{amsfonts} 
\usepackage{textcomp}
\usepackage{arydshln}
\usepackage{stmaryrd}
\usepackage{mathrsfs}
\usepackage{booktabs}
\usepackage{balance}
\usepackage{multirow}
\usepackage{overpic}
\usepackage{cite}
\usepackage[font=small,labelfont=bf]{caption}
\usepackage{color}
\usepackage{pifont}
\usepackage{tikz}
\usetikzlibrary{patterns}

\usepackage{colortbl}
\usepackage{pgfplots}
\pgfplotsset{compat=1.17}
\usepackage{tabularx}
\usepackage{arydshln}
\usepackage{amssymb}
\usepackage{pifont}

 \usepackage{amsmath}
\usepackage{wasysym}%
\usepackage{enumitem}
\usepackage{wrapfig}
\usepackage{xspace}
\usepackage{tabu}
\usepackage[super]{nth}
\usepackage{subcaption}
\usepackage{scalefnt}
\usepackage{dsfont}
\usepackage{graphicx,calc}

\definecolor{darkgreen}{RGB}{0,153,51}
\definecolor{linkgreen}{RGB}{52,130,48}
\definecolor{LightCyan}{rgb}{0.87,0.92,0.96}
\definecolor{m_green}{RGB}{233, 254, 187}
\definecolor{m_orange}{RGB}{255, 212, 121}
\definecolor{m_red}{RGB}{255, 190, 188}
\definecolor{m_violet}{RGB}{215, 131, 255}
\definecolor{m_blue}{RGB}{186, 234, 255}
\definecolor{m_brown}{RGB}{255,212,120}
\definecolor{m_lightgreen}{RGB}{212,251,122}
\definecolor{notetext}{rgb}{0.7,0,0}

\definecolor{model_pink}{RGB}{235, 106, 164}
\definecolor{model_orange}{RGB}{250, 194, 122}
\definecolor{model_green}{RGB}{164, 210, 162}
\definecolor{model_gray}{RGB}{120, 120, 120}
\definecolor{model_yellow}{RGB}{251, 231, 171}
\definecolor{model_purple}{RGB}{190, 146, 211}

\newcommand{\name}{SceneGraphLoc}

\newcommand{\xmark}{\ding{55}}%

\def\eg{\emph{e.g.}\@\xspace} 
\def\ie{\emph{i.e.}\@\xspace}

\renewcommand{\vec}[1]{\ensuremath{\mathbf{#1}}}

\begin{document}

\title{\name{}: Cross-Modal Coarse Visual Localization on 3D Scene Graphs}
\author{Yang Miao$^{1}$, Francis Engelmann$^{1,2}$,  Olga Vysotska$^{1}$, Federico Tombari$^{2,3}$, 
Marc Pollefeys$^{1,4}$, Dániel Béla Baráth$^{1}$}
\institute{$^{1}$ETH Zurich, $^{2}$Google, $^{3}$TU Munich, $^{4}$Microsoft}
\maketitle
\vspace*{-5mm}
\begin{center}
    \centering
        \includegraphics[width=0.99\columnwidth]{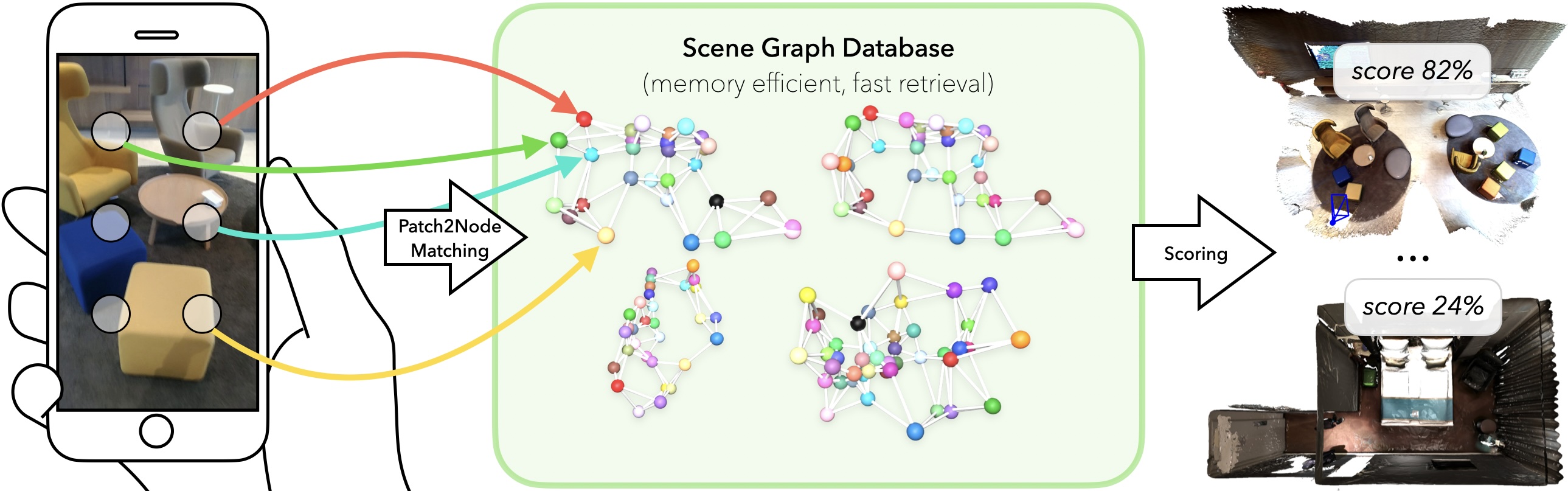}
        \captionof{figure}{\textbf{SceneGraphLoc} addresses the novel problem of localizing a query image in a database of 3D scenes represented as compact multi-modal 3D scene graphs.}
        \label{fig:teaser}
\end{center}

\begin{abstract}
    This paper introduces a novel problem, \ie, the localization of an input image within a  multi-modal reference map represented by a database of 3D scene graphs. 
    These graphs comprise multiple modalities, including object-level point clouds, images, attributes, and relationships between objects, offering a lightweight and efficient alternative to conventional methods that rely on extensive image databases. 
    Given the available modalities, the proposed method \name{} learns a fixed-sized embedding for each node (\ie, representing an object instance) in the scene graph, enabling effective matching with the objects visible in the input query image.
    This strategy significantly outperforms other cross-modal methods, even without incorporating images into the map embeddings.
    When images are leveraged, \name{} achieves performance close to that of state-of-the-art techniques depending on large image databases, while requiring \textit{three} orders-of-magnitude less storage and operating orders-of-magnitude faster. 
    The code is public: \href{https://scenegraphloc.github.io/}{scenegraphloc.github.io}.
\end{abstract}

\section{Introduction}

Coarse visual localization, or place recognition, is a fundamental component in computer vision and robotics applications, defined as the task of identifying the approximate location where a query image was taken, given a certain tolerance level 
\cite{liu2019stochastic,Torii2018,Doan2019,arandjelovic2016netvlad,Hausler2021,Zaffar2021,Torii2021,Garg2019,Hausler2019,Khaliq2020,Berton2021,Ibrahimi2021,Warburg2020}. 
This capability is crucial for estimating the state of robots and is widely utilized in autonomous, unmanned aerial, terrestrial, and underwater vehicles, as well as AR/VR devices. 
The task is typically approached as an image retrieval problem, where the image to be localized is compared against a large database of posed images and, optionally, a 3D reconstruction of the scene. 
The most similar images retrieved from the database are used to estimate the precise location of the query image.

The challenge with current state-of-the-art image-based coarse localization methods, such as~\cite{keetha2023anyloc}, is their dependency on extensive image databases, which are not only \textit{storage-heavy} but also \textit{slow} to query, despite optimizations through hashing and other tricks. 
Moreover, these methods typically necessitate that the query and database share the same modality, limiting the scope of their application. 
Cross-modal approaches, such as~\cite{zhou2022geometry,sarlin2023orienternet}, which attempt to bridge different types of data, often restrict their scope to connecting two modalities at a time (\eg, image-to-point cloud or image-to-bird's eye view map), one for the query and one for the database, thus narrowing their potential applications.

\begin{figure}[t] 
\begin{center}
\includegraphics[width=0.99\columnwidth]{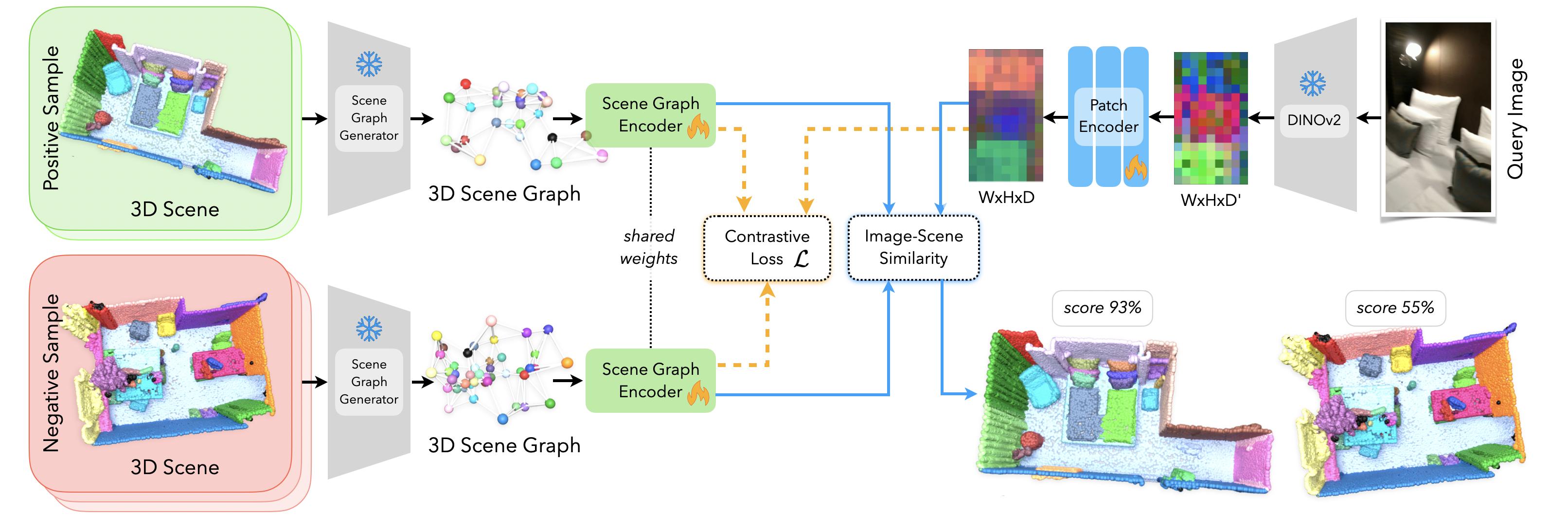}
    \caption{\textbf{Overview.} The training phase is represented by orange arrows, while blue arrows denote the inference phase. 
    During training, a query image and its associated 3D scene graph form a positive sample within a contrastive learning framework, where negative samples are generated by associating scene graphs of different scenes with the same query image. The objective is to learn the embeddings of both the graph and the image so that embeddings of the positive pair are drawn closer, whereas those of the negative pair are pushed apart. 
    In the inference phase, the task involves assigning the correct scene graph to a given query image from a selection of multiple graphs, achieved by optimizing the cosine similarity between their embeddings. 
    } \label{fig:pipeline}
\end{center}
\end{figure}  

This paper addresses the \textit{novel} challenge of localizing a query image within a database that is represented not by conventional images but by the 3D scene graph~\cite{armeni20193d, wald2020learning} that integrates a diverse set of modalities, including point clouds, images, semantics, object attributes, and relationships. 
We tackle this problem by learning to map these modalities into a unified embedding space, thus allowing us to represent indoor scenes compactly through their objects (\eg, table and wall). 
This method enables the creation of small, efficient databases and significantly accelerates the coarse localization process.

\noindent
\textbf{Contributions.} The primary contributions of this paper are as follows:

\begin{enumerate}
    \item Introducing a novel problem: cross-modal localization of a query image within 3D scene graphs incorporating a mixture of modalities.
    \item SceneGraphLoc, a new method for the coarse localization of an input image given a reference map represented by a database of 3D scene graphs. 
\end{enumerate}

\noindent
Even without incorporating images into the map, SceneGraphLoc largely outperforms other cross-modal methods on two large-scale, real-world indoor datasets.
With images, SceneGraphLoc achieves performance close to that of state-of-the-art image-based methods while requiring \textit{three} orders-of-magnitude less storage and operating orders-of-magnitude faster. 
The method is visualized in Fig.~\ref{fig:teaser}.
\section{Related Work}


\textbf{Localization}, the process of determining the position and orientation of an agent within a pre-built map, is pivotal across various domains such as mobile robots~\cite{heng2019project, lim2012real, zhao2023stablesafebarrier, zhao2024stablesafehumanalignedreinforcement}, and augmented reality~\cite{castle2008video,lynen2020large}.
The differentiation in localization techniques arises from their scene representation methods -- be it through explicit 3D models~\cite{Germain2019,Germain2020,Irschara2009,Li2012,Liu2017,Lynen2020,Sarlin2019,sarlin2021back,Sattler2017,schonberger2018semantic,Svarm2017,zeisl2015camera,SarlinSuperGlue,zhang2023revisitingrotationaveraginguncertainties }, sets of posed images~\cite{Zhang2006,Zheng2015,pion2020benchmarking,bhayani2021calibrated}, or implicitly via neural network weights~\cite{Kendall2015,Valentin2015,Walch2017,Kendall2017, Brachmann2017, Brachmann2018,Balntas2018,Cavallari2019,brachmann2021visual,Moreau2021} -- and their approach to camera pose estimation, whether by 2D-3D~\cite{Germain2019,Germain2020,Sarlin2019,SarlinSuperGlue,Sattler2017,Svarm2017,zeisl2015camera} or 2D-2D~\cite{Zheng2015,bhayani2021calibrated} matches, or through a composite of base poses~\cite{Kendall2015,Walch2017,Kendall2017,sattler2019understanding,pion2020benchmarking,Moreau2021}. 
In practice, localization comprises two main steps: a coarse and precise stage. 
Here, we focus on the coarse step, finding potential locations of a query image.

\noindent
\textbf{Coarse Localization} (or place recognition) is often cast as an image retrieval problem~\cite{GargPlaceRecognition,arandjelovic2016netvlad,Berton2021,Hausler2021,Ibrahimi2021,Kim2017,Peng2021_ICRA,Peng2021_ICCV,Warburg2020,berton2022rethinking,keetha2023anyloc} that consists of two phases. 
In the offline indexing phase, a reference map (image or point cloud database) is gathered. 
In the online retrieval phase, a query image -- captured during a future traverse of the environment -- is localized coarsely by retrieving the closest match to this image in the reference map. 
Recent methods perform the retrieval using learned embeddings that are produced by a feature extraction backbone equipped with a head that implements some form of aggregation or pooling, the most notable being NetVLAD~\cite{arandjelovic2016netvlad}.
While these methods achieve impressive results, they are limited to a single modality (\eg, images) and require large databases.

\noindent
\textbf{Localization using multi-modal data}.
While dense mesh models are not as widely adopted as sparse Structure-from-Motion-based approaches, they have nonetheless been the focus of considerable research efforts~\cite{panek2022meshloc,Aubry2014,Aubry2016,Brejcha2020,Grelsson2020,Mueller2019,Plotz2017,Ramalingam2010,Shan2014,Sibbing2013,Tomesek2022,Cadik2018,Zhang2020,zhou2022geometry}. 
The body of prior work can be broadly segmented into two main strategies: 
The first entails the precise alignment of actual images with three-dimensional models (which may be coarse) through applying specialized techniques such as ones using contours~\cite{Plotz2017} or skylines~\cite{Ramalingam2010}. 
The second strategy emphasizes the identification and matching of local image features~\cite{Aubry2014,Aubry2016,Brejcha2020,Mueller2019,Grelsson2020,Shan2014,Sibbing2013,Tomesek2022,Zhang2020,zhou2022geometry,wang2023dgc}, a method that has gained traction for its ability to match real-world images with non-photorealistic renderings of colored meshes, or even meshes without color~\cite{Brejcha2020,panek2022meshloc,Tomesek2022}.
CAD and other models are also commonly used by object pose estimation~\cite{Aubry2014,Ponimatkin2022,Labbe2020,Hodan2015,Hodan2020,Hodan2021,Gumeli2021,Grabner2018,Georgakis2019}. 
%
Image to LiDAR localization~\cite{bernreiter2021spherical,gao2023visual,shubodh2024lip,hess2024lidarclip} is also relevant, especially in robotics applications.

Another variant of multi-modal data localization involves cross-view matching. This technique determines the camera position by finding correspondences between a ground-level query image and a two-dimensional bird’s eye view map, such as a satellite image or a semantic landscape map \cite{Shan2014,Viswanathan2014,Lin2015,Workman2015,Hu2018,Hu2019,sarlin2023orienternet}. 
Other cross-modal techniques were also proposed to involve semantics~\cite{garg2020semantics,Garg2019} and event cameras~\cite{ji2023cross} in image-based localization.

These approaches, while demonstrating promising localization results, are limited to interactions between two modalities -- one for the query and one for the reference database. 
Our approach, in contrast, seeks the cross-modal coarse localization of a query image in a database composed of multiple modalities, \eg, 3D point clouds, images, semantics, object attributes, and relationships.


\noindent
\textbf{Scene representation}, encapsulating various scene attributes has evolved significantly, yielding diverse surface representations from explicit forms (3D point clouds~\cite{qi2017pointnet, miao2021tianjiport}, meshes~\cite{hanocka2020point2mesh}, surfels~\cite{stuckler2014multi}) to implicit ones (occupancy~\cite{kutulakos2000theory}, signed distance functions~\cite{curless1996volumetric,izadi2011kinectfusion}). 
The advent of neural representations has introduced novel means of encoding geometry~\cite{mescheder2019occupancy,park2019deepsdf,peng2020convolutional,weder2021neuralfusion}, appearance~\cite{oechsle2019texture,mihajlovic2021deepsurfels}, and semantics~\cite{murez2020atlas, miao2024vscp3d}. 
A comprehensive review is provided by Tewari et al.~\cite{tewari2022advances}. 
The integration of directions/rays~\cite{mildenhall2021nerf} and visibility encoding in surface reconstruction~\cite{savinov2016semantic} has further enriched this domain. Armeni et al. \cite{armeni20193d} introduced the 3D scene graph structure as a multi-layer representation of a scene that captures geometry, semantics, objects, and camera poses in a unified manner. Subsequent efforts \cite{wald2020learning,rosinol20203d} have further advanced 3D scene graph learning and structure. 

The increasing interest in 3D scene graphs~\cite{armeni20193d,rosinol20203d,wald2020learning,kim20193, Kabalar2023longtermloc}
underscores their potential as structured, rich descriptors for real-world scenes. 
Methods range from online incremental construction~\cite{wu2021scenegraphfusion,hughes2022hydra} to offline generation from RGB-D imagery~\cite{armeni20193d,wald2020learning,Rosinol21}, and approaches for scene graph prediction~\cite{Zhang21a,Zhang21b}. 
Their application spans embodied AI~\cite{Sepulveda18,Rosinol21,Ravichandran22}, task completion~\cite{gadre2022continuous,agia2022taskography}, variability estimation~\cite{Looper23}, and SLAM~\cite{Rosinol21,hughes2022hydra}. 
Recent studies like~\cite{Ying23} introduce frameworks for localizing unseen objects by utilizing 3D scene graphs and graph neural networks for relation prediction, showcasing the utility of scene graphs in enhancing spatial understanding. 
Similarly, \cite{sarkar2023sgaligner} offers new perspectives on 3D scene alignment, employing node matching within overlapping scene graphs to facilitate precise 3D map alignment. 
Kabalar et al.~\cite{Kabalar2023longtermloc} assume that the query image has been coarsely localized and leverages a scene graph to identify dynamic objects and for precise localization with image features. 
%
Despite these significant advancements underscoring the value of scene graphs, their potential in multi-modal localization remains largely untapped.
In this paper, we use a scene graph representation of the map in which we aim to localize a query image. 

\section{Visual Localization with 3D Scene Graph}


\noindent
\textbf{Problem Statement.}
Let us assume that we are provided with a pre-constructed map of the environment, denoted as $\mathcal{G}$, which is represented as a set of $N \in \mathbb{N}^+$ 3D scene graphs $\mathcal{G}_i = (\mathcal{V}_i, \mathcal{E}_i)$, such that $\mathcal{G} = \{ \mathcal{G}_i \}_{i \in [0,N)}$. 
Having separate graphs $\mathcal{G}_i$ is analogous to the hierarchy levels presented in~\cite{Rosinol21,hughes2022hydra}, each representing a group of object instances constituting a place like a room or building.
Moreover, this task can also be straightforwardly redefined as subgraph selection. 
Vertices $v \in \mathcal{V}_i$ symbolize instances of objects (\eg, chairs, tables) and large instances of semantic categories (\eg, walls, ground) within the scene, $i \in [0, N)$. 
Let $\mathcal{V} = \bigcup_{i \in [0, N)} \mathcal{V}_i$ aggregate all objects across the scenes.
Edges $\mathcal{E}_i = \{ (v^i_j, v^i_k) | v^i_j, v^i_k \in \mathcal{V}_i \}$ delineate the relationships between objects, such as ``nearby'', ``standing on'', and ``attached to''.

For each vertex $v$, we introduce $M \in \mathbb{N}^+$ map modalities $f_j: \mathcal{V} \rightarrow \mathcal{M}_j$ for $j \in [0, M)$, where $f_j$ maps vertex $v$ to the $j^{th}$ modality $\mathcal{M}_j$ that may be $\mathcal{M}_j \in \{ \texttt{position}, \; \texttt{orientation}, \; \texttt{point cloud}, \; \texttt{semantic category}, \texttt{image}, \texttt{attribute} \}$. 
While this paper focuses on these modalities, the set is easily extendable by incorporating additional ones, such as \texttt{textual description} or \texttt{floor plan}.

Given $\mathcal{G}$ and an input query image $I$, the objective is to identify the scene graph $\mathcal{G}_i$ corresponding to the space depicted in image $I$. 
Note that while we focus on having an input image in this paper, the method can be modified to other modalities as well, \eg, depth image. 
Formally, we aim to resolve problem:
\begin{equation}
    \mathcal{G}_{i^*} = \underset{i \in [0, N)}{\arg\max} \, \llbracket \texttt{contains}(\mathcal{G}_i, \mathbf{p}_I) \rrbracket,
\end{equation}
where $\mathbf{p}_I \in \mathbb{R}^3$ is the unknown 3D position of the image, and $ \llbracket . \rrbracket$ is the Iverson bracket which equals to $1$ if the condition inside holds and $0$ otherwise.
It is important to note that our objective is \textit{coarse} localization, opting for the selection of $\mathcal{G}_i$ without needing precise estimation of $\mathbf{p}_I$.

To this end, the optimization problem can be reformulated to incorporate the chirality constraint, asserting that if an object is visible in image $I$, it must be positioned in front of the camera in 3D space, unobstructed by any entities (\eg, walls). 
Consequently, the problem becomes:
\begin{equation}
    \mathcal{G}_{i^*} = \underset{i \in [0, N)}{\arg\max} \sum_{o_I \in \mathcal{O}_I} \left\llbracket \texttt{visible}(\mathcal{G}_i, o_I) \right\rrbracket \approx 
    \underset{i \in [0, N)}{\arg\max} \sum_{o_I \in \mathcal{O}_I} \log P(o_I \; | \; \mathcal{G}_i),
    \label{eq:optimization}
\end{equation}
where $\mathcal{O}_I$ is the object set visible in image $I$, 
object $o_I \subseteq \{ (x, y) \in I \}$ is a set of pixels in the image ($o_I \in \mathcal{O}_I$),
$\texttt{visible} : \mathcal{G}_i \times \mathcal{O}_I \to \{ 0, 1 \}$ indicates if an object is visible in a scene graph, 
$P(o_I \; | \; \mathcal{G}_i)$ is the probability of $o_I$ stemming from $\mathcal{G}_i$.

Function $\texttt{visible}$ can be approached as an indicator of whether object $o_I$ appears in graph $G_i$. It holds if and only if there exists a $v \in \mathcal{V}_i$ such that $v$ represents the same object as $o_I$.
Therefore, this problem can be approached as matching a set of image pixels ($o_I$) to a scene graph node ($v$) that comprises a set of modalities.
In the next sections, we will describe a way to learn a unified embedding space for both $o_I$ and $v$ such that they become matchable. 

\noindent
\textbf{Proposed Pipeline}
is shown in Fig.~\ref{fig:pipeline}. 
It consists of two concurrent stages:
the first one generates object embeddings $e_q \in \mathbb{R}^D$ from patches $q \in \mathcal{Q}_I$ within the query image $I$, and the second derives node embeddings $e_v \in \mathbb{R}^D$ for nodes $v \in \mathcal{V}_i$ in the scene graph $\mathcal{G}_i$. 
The training objective is to make $\delta(e_q, e_v) = 0$ if and only if the object associated with node $v$ is directly visible (\ie, neither occluded nor outside the camera frustum) through the image patch $q$. 
Here, $\delta: \mathbb{R}^D \times \mathbb{R}^D \to \mathbb{R}$ denotes inverse cosine similarity normalized to $\left[0,1\right]$. 

After generating $e_q$ and $e_v$, the model performs nearest neighbor matching (NN) for each patch in each scene graph $\mathcal{G}_i$, assigning node $v$ to $q$ such that
\begin{equation} \label{eq: patch_association}
    \text{NN}(q, \mathcal{V}_i) = \underset{v \in \mathcal{V}_i}{\arg\min} \; \delta(e_q, e_v).
\end{equation}
Through matching, we establish patch-to-node correspondences $\mathcal{C}_i = \{ (q, v) \; | \; v = \text{NN}(q, \mathcal{V}_i) \in \mathcal{V}_i, q \in \mathcal{Q}_I \}$.
Based on $\mathcal{C}_i$, we devise an image-to-graph similarity score enabling us to deduce whether image $I$ corresponds to the space represented by $\mathcal{G}_i$.
Finally, potential coarse locations of $I$ are selected by maximizing the similarity score across the stored scene graphs as depicted in Eq~\ref{eq:optimization}.

\subsection{Object Embeddings in the Scene Graph}
\label{sec:graphencoder}
\begin{wrapfigure}{r}{0.40\textwidth}
\vspace{-30px}
  \begin{center}
    \includegraphics[width=0.40\textwidth]{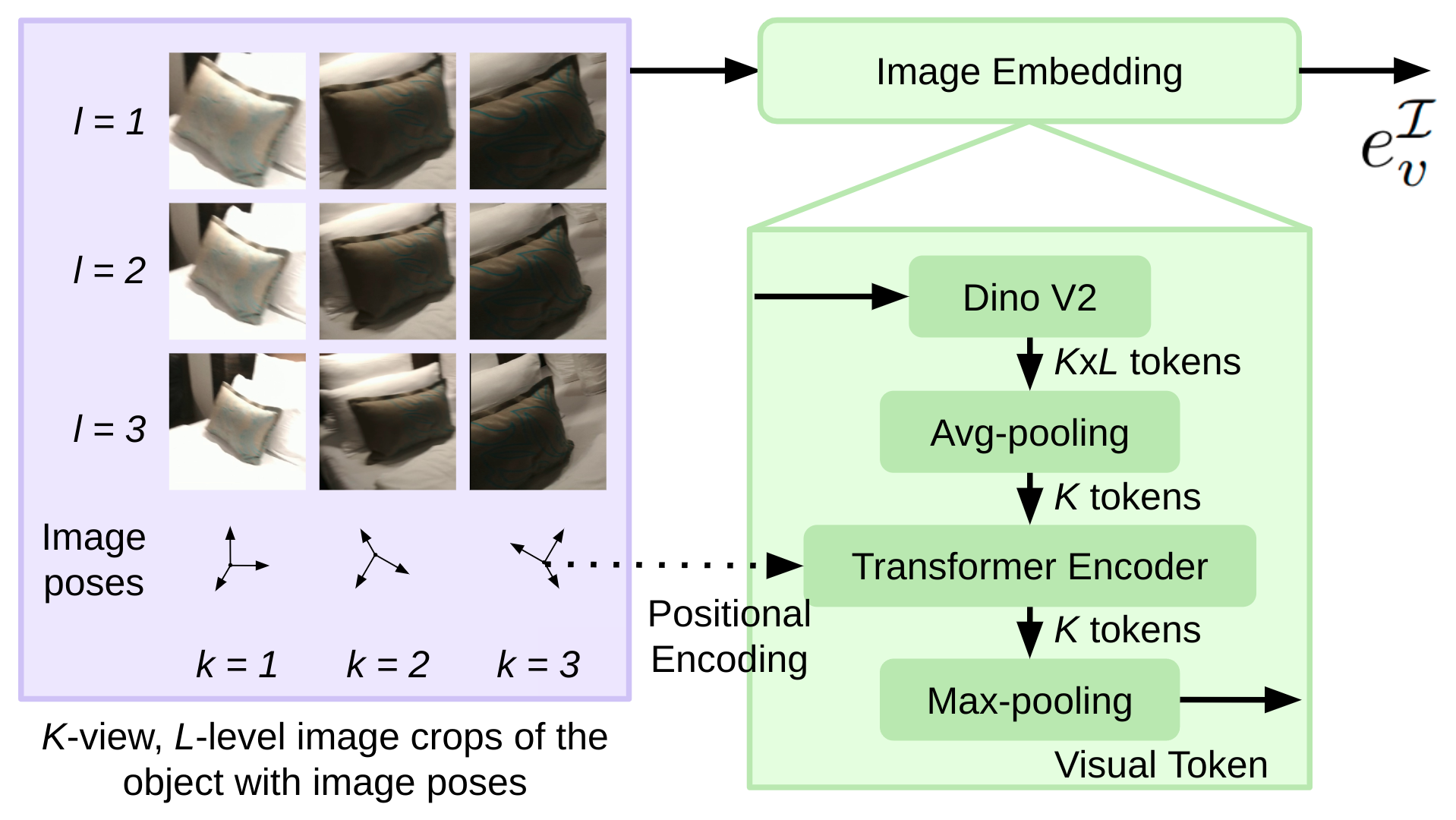}
  \end{center}
\vspace{-10px}
  \caption{The embedding of image modality $\mathcal{I}$ for each object. The image crops of a pillow are shown as an example.}
  \label{fig:scenegraph_encoder}
\vspace{-25px}
\end{wrapfigure}

This section aims to obtain an embedding for each node within graph $\mathcal{G}_i$, encapsulating information from \textit{all} available modalities. 
Our method builds upon the method of Sarkar et al.~\cite{sarkar2023sgaligner}, with enhancements to include the \texttt{image} modality and to distill a unified embedding from all modalities rather than merely concatenating separate embeddings as in~\cite{sarkar2023sgaligner}.

Scene graphs are conceptualized as multi-modal knowledge graphs, similar to those used in entity alignment, treating semantic and geometric information as distinct modalities. 
The objective is to learn a joint multi-modal embedding from the individual modal encodings (uni-modal), ensuring nodes corresponding to the same object instance across different graphs are closely positioned. 
This involves the creation of uni-modal embeddings for the three primary types of 3D scene graph information: 
\textit{object} embeddings encoding nodes in $\mathcal{V}$, 
\textit{structure} embeddings $\mathcal{S}$ representing edges in $\mathcal{E}$ as a structured graph, 
and two \textit{meta} modalities encoding attributes ($\mathcal{A}$) and relationships ($\mathcal{R}$) between objects as one-hot vectors. 
These uni-modal embeddings are then combined in a weighted manner and optimized jointly through knowledge distillation.

Each of these modalities is processed separately to generate uni-modal embeddings, which are subsequently integrated to model complex inter-modal interactions within the joint embedding space.

\par \noindent \textbf{Object Embedding.} 
Node $v \in \mathcal{V}$ may contain multiple modalities, such as \texttt{point cloud} ($\mathcal{P}$) and \texttt{image} ($\mathcal{I}$).
Point clouds contain rich geometric information about objects. 
The point cloud corresponding to each $v \in \mathcal{V}$ is inputted to the object encoder. 
We employ the PointNet architecture \cite{qi2017pointnet} as the object encoder to extract the geometric feature $e^{\mathcal{P}}_v$ for every node.
Furthermore, we integrate multi-level and multi-view visual embeddings to enrich the graph encoder with a more nuanced understanding of image information. 
The visual embedding pipeline is visualized in Fig.~\ref{fig:scenegraph_encoder}.
%
%
For each node $v$ denoting a 3D object, the top $K_{view} \in \mathbb{N}^+$ images with largest visibility of $v$ is selected: $\{ I_{\text{db},k}^v \; | \; k = [0, K) \} \subseteq \mathcal{I}_\text{db}$ from the database $\mathcal{I}_\text{db}$. 
We can define an ordering over these images such that $\phi(I_{\text{db},0}^v, v) \geq \phi(I_{\text{db},1}^v, v) \geq ... \geq \phi(I_{\text{db},K_{view} - 1}^v, v)$, with the visibility function $\phi$ quantifying the extent of node $v$ observed in each image through pixel count.
Visibility check is implemented by projecting the 3D model of $v$ to each posed image in $\mathcal{I}_\text{db}$, which are usually available when constructing the scene graph~\cite{wu2021scenegraphfusion, hughes2022hydra}. 
Parameter $K_{view} = 10$ in all our experiments. 

Drawing inspiration from OpenMask3D~\cite{takmaz2023openmask3d}, for any given view $I^{v}_\text{db}$ of object $v$, initial steps include calculating the bounding box ${b}_{v,0}$ of $v$ within the image, followed by the generation of multi-level bounding boxes $\{{b}_{v,l} \; | \; l \in [0, L) \}$ through iterative enlargement of ${b_{v,0}}$. 
This enlargement strategy aims to capture contextual information around object $v$.
Subsequently, Dino V2~\cite{oquab2023dinov2} processes the image crops defined by ${{b}_{v,l}}$, extracting multi-level features $\{ {f}_l \; | \; l \in [0, L) \}$. 
For each image $I_{\text{db},k}^v$, an average pooling operation aggregates these multi-level crop features into a singular feature vector $f = \text{avg pool} \{{f}_l \; | \; l \in [0, L) \}$.
The final step of this process involves the application of a Transformer encoder, which incorporates image poses as positional encodings. 
This step synthesizes multi-view object tokens into a cohesive visual embedding $e^{\mathcal{I}}_v$, effectively integrating the diverse perspectives and levels of contextual information pertaining to each object within the database.
Please note that the image database does not necessarily have to be stored after distilling the object embeddings.

\noindent \textbf{Structure Embedding.} 
3D Scene Graphs encapsulate the object relationships, which we exploit to encode their spatial configuration. 
This relational data is represented through a \textit{structure} graph, where node features embody the relative translations between object instances, and edges denote these relationships. 
The relative translation is determined by calculating the distance from the object instance to any other object in the scene. 
To encapsulate this structural information within $\mathcal{G}_i$, a Graph Attention Network (GAT)~\cite{veličković2018graph} is utilized, with the weight matrix constrained to a diagonal form to reduce computational demands and enhance model scalability. 
Following the method in \cite{sarkar2023sgaligner}, the structural embedding $e^{\mathcal{S}}_v$ is derived from the final layer of a two-layer GAT model. 

\noindent \textbf{Meta Embeddings.}
In addition to geometry and structure, the object attributes and the inter-object relationships are captured in two distinct embeddings, $e^{\mathcal{R}}_v$ and $e^{\mathcal{A}}_v$. 
The relationships an object maintains with others are conceptualized as a bag-of-words~\cite{csurka2004visual} feature vector, which is input to a feed-forward neural network, distilled in relational embedding $e^{\mathcal{R}}_v$. 
A similar approach is employed for the attributes associated with the objects, producing embedding $e^{\mathcal{A}}_v$. 

\noindent \textbf{Joint Embedding.} 
Similar to~\cite{sarkar2023sgaligner}, we concatenate uni-modal features to a compact representation for each object $v$.
Contrary to~\cite{sarkar2023sgaligner}, we encapsulate it in a multi-layer perceptron ($\text{MLP}$) to learn an embedding with size independent of the available modalities, fusing information from all. 
Embedding $e_v$ is as:
\begin{equation}
    e_v = \text{MLP}\left(\oplus_{k \in \mathcal{K}}\left[\frac{\exp(w_k)}{\sum_{j \in \mathcal{K}} \exp(w_j)} e^k_v \right] \right),    
\end{equation}

\noindent where $\oplus$ is the concatenation operator, $\mathcal{K} = \{\mathcal{P}, \mathcal{I}, \mathcal{S}, \mathcal{R}, \mathcal{A} \}$, and $w_m$ is a trainable attention weight for each modality $k \in \mathcal{K}$.
A two-layer MLP is applied to map the dimension of the concatenated multi-modal descriptors from $D^{\mathcal{K}}$ to $D$.
We apply $L_2$ normalization to each uni-modal feature before concatenation. 

\noindent \textbf{In practice},
these independent modalities are only required and used in the mapping phase of the procedure.
This stage involves the construction of an environmental map in the form of a 3D scene graph, during which each node $v$ is distilled into an embedding $e_v$. 
During localization, we can ignore independent modalities and use only the fixed-sized embeddings $e_v$. 

\subsection{Object Embeddings in the Query Image}

To solve the optimization problem in Eq.~\ref{eq:optimization}, we need to find object instances $o_I \in \mathcal{O}_I$ in query image $I$.
A straightforward approach to do so would be to apply a 2D panoptic segmentation algorithm, \eg, ~\cite{cheng2021mask2former}. 
However, we noticed in our experiments that such an approach is susceptible to inaccuracies and failures (over- and under-segmentation) in the segmentation, severely affecting the accuracy. 
Thus, we approach the problem by a visual Transformer (ViT), breaking up the image into rectangular patches $q \in \mathcal{Q}_I$ and distilling an independent embedding for each patch $q$ based on the object visible in $q$. 
We use Dino V2~\cite{oquab2023dinov2} as a backbone to obtain patch-level features.
These features are passed through a patch encoder trained to create embeddings from the patch features considering the objects visible from each $q$.
For this encoder, we use a 4-layer convolution neural network (CNN) with residual blocks introduced in \cite{resnet} on the Dino V2 features and a 3-layer MLP to further map the patch feature to dimension $D$.


\subsection{Contrastive Learning}
\begin{wrapfigure}{r}{0.33\textwidth}
\vspace{-60px}
  \begin{center}
    \includegraphics[width=0.33\textwidth]{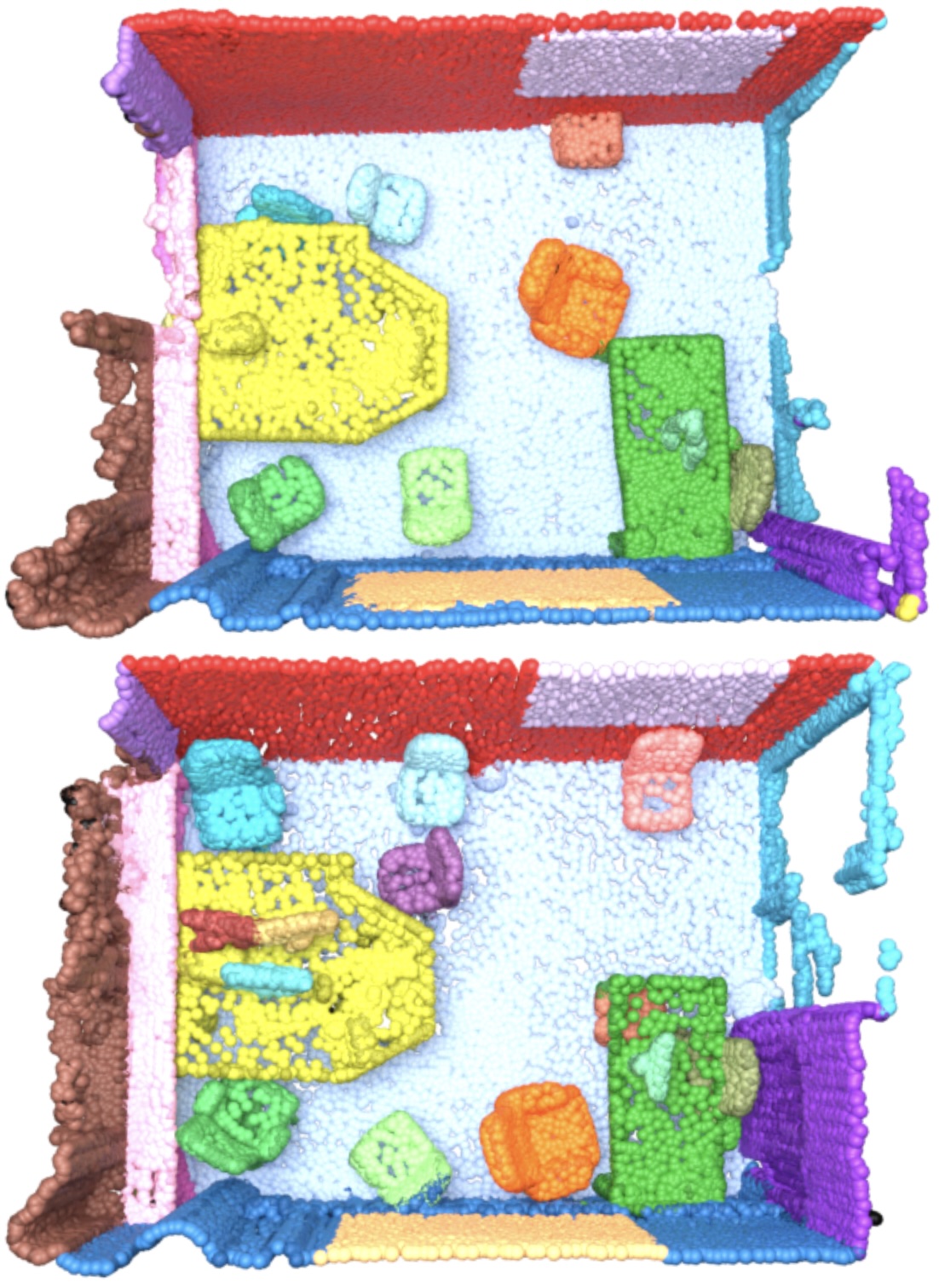}
  \end{center}
\vspace{-10px}
  \caption{The same scene at different points in time from the 3RScan dataset~\cite{3rscan}.}
  \label{fig:temporal_changes}
\vspace{-45px}
\end{wrapfigure}

We use contrastive learning to learn a joint embedding space for the scene graph nodes and image patches. 
To do so, we form query image and graph pairs $(I, \mathcal{G}_I)$. 
Real-world scenes are rarely static, \eg, objects move or undergo non-rigid deformations and illumination changes \cite{3rscan}.
To ensure that the learned embedding is robust to such temporal changes,
we use scene graph $\mathcal{G}_I$ from the same temporal point when $I$ was captured, as positive samples, as well as a scene graph $\mathcal{G}^t_I$ from another scan. 
Graph $\mathcal{G}^t_I$ represents the same place as $\mathcal{G}_I$, but it undergoes temporal changes.
An example is shown in Fig.~\ref{fig:temporal_changes}. 
\newpage
For a query image $I$, a set of candidate graphs $\{\mathcal{G}_I, \mathcal{G}^t_I, \mathcal{G}_1, ..., \mathcal{G}_N \}$ is provided for training, where $\{\mathcal{G}_1, ..., \mathcal{G}_N \}$ act as $N$ negative samples, depicting different scenes than the target scene of the query image. 
We train our model by optimizing both a static loss and a temporal loss as:
\begin{equation} \label{eq:train_loss}
    \mathcal{L} = \alpha * \mathcal{L}_\text{static} + (1-\alpha) * \mathcal{L}_\text{temp}.
\end{equation}
During training, we assume that image patch to graph node pairs are available\cite{3rscan} such that $P_I = \{(q, v) \; \vert \; q \in \mathcal{Q}_I, v \in \mathcal{G}_I\}$ and $P_I^t = \{(q, v^t) \; \vert \; q \in \mathcal{Q}_I, v \in \mathcal{G}^t_I \}$. 
For each pair $(q, v)$ from $P_I$ and each pair from $P_I^t$, we use the following notation.
    Set $N^\mathcal{I}_q = \{ q' \; \vert \; q' \in \mathcal{Q}_I, v_{q'} \neq v \}$ contains patches seeing objects other than $v$.
    $N^\mathcal{G}_v = \{ v_n \; \vert \; v_n \in \mathcal{V}_I\cup\mathcal{V}_1\cup...\cup\mathcal{V}_N  \} \setminus \{ v \}$ contains the 3D objects of all candidate scene graphs other than $v$, where $\mathcal{V}_I$ represents the objects nodes of the target graph $\mathcal{G}_I$ and $\mathcal{V}_i$ is the nodes of other graphs $\mathcal{G}_i$.
    $N^{\mathcal{G}^t}_v = \{ v_n \vert v_n \in \mathcal{V}^t_I\cup\mathcal{V}_1\cup...\cup\mathcal{V}_n \} \setminus \{ v \}$, where $\mathcal{V}^t_I$ represent the nodes of the target graph $\mathcal{G}^t_I$.
%
The static loss is defined as bi-directional N-pair loss~\cite{lin2022multi} as follows:
\begin{equation} \label{eq:train_loss_static}
    \mathcal{L}_\text{static} = E_{P_I \in B} \left[ E_{(q,v) \in P_I} \left[ -\log\bigg(\frac{1}{2}p(q, v, N^\mathcal{I}_q, N^\mathcal{G}_v) + \frac{1}{2}p(v, q, N^\mathcal{I}_q, N^\mathcal{G}_v)\bigg) \right] 
 \right], 
\end{equation}
where $E_{P_I \in B}$ represent loss averaging over a batch of query images and their corresponding candidate scene graphs; 
$p(q, v, N^\mathcal{I}_q, N^\mathcal{G}_v)$ and $p(v, q, N^\mathcal{I}_q, N^\mathcal{G}_v)$ represent the bi-directional probability distributions of the positive pair as:
\begin{equation*} \label{eq:pairs_prob_distributions}
    \begin{split}
        p(q, v, N^\mathcal{I}_q, N^\mathcal{G}_v) &= \frac{ f(e_q, e_v)}{f(e_q, e_v) + 
                  \sum_{q_n \in N^\mathcal{I}_q}{f(e_q, e_{q_n})} +
                  \sum_{v_n \in N^\mathcal{G}_v}{f(e_q, e_{v_n})}}
                 , \\
    \end{split} 
\end{equation*}
where $f(e_q, e_v) = \text{exp}(\frac{-\delta(e_q, e_v)}{\tau})$, $\delta(e_q, e_v)$ represents the inverse cosine similarity between embeddings $e_q$ and $e_v$, and $\tau$ is a temperature parameter. Probability distribution $p(v, q, N^\mathcal{I}_q, N^\mathcal{G}_v)$ is written similarly. 
The temporal loss term is defined the same as Eq.\ref{eq:train_loss_static} but with $P^t_I$ and $N^{\mathcal{G}^t}_v$ as follows:
\begin{equation} \label{eq:train_loss_temporal}
    \mathcal{L}_\text{temp} = E_{P^t_I \in B} \left[ E_{(q,v) \in P^t_I} \left[ -\log\bigg(\frac{1}{2}p(q, v, N^\mathcal{I}_q, N^{\mathcal{G}^t}_v) + \frac{1}{2}p(v, q, N^\mathcal{I}_q, N^{\mathcal{G}^t}_v) \bigg) \right] 
 \right]. 
\end{equation}
By minimizing Eq.\ref{eq:train_loss}, the paired cross-modal embeddings $e_q$ and $e_v$ are pulled together while the embeddings from different objects are pushed apart.

\subsection{Scene Graph Retrieval}

Given a pre-established map of an environment represented through a collection of scene graphs $\mathcal{G} = \{ \mathcal{G}_i \; | \; i \in [0,N) \}$, where each node embedding has been precomputed, the goal during inference is to identify the top-$K$ scene graphs $\mathcal{G}_i$ in which image $I$ was likely captured. 
The method to address this challenge involves calculating the similarity between a graph and the image as follows:
\begin{equation} \label{eq: scene_score}
    s(\mathcal{G}_i, I) = \frac{1}{|\mathcal{Q}_I|} \sum_{q \in \mathcal{Q}_I} \left[1 -\delta(e_q, \text{NN}(q)) \right],
\end{equation}
where $\mathcal{Q}_I$ denotes the set of image patches in $I$, and for each patch $q$, $\text{NN}(q, \mathcal{V}_i) \in \mathcal{V}_i$ represents the nearest node in terms of embedding similarity. 
The function $s$, which is assumed to map values to the interval $[0, 1]$, facilitating the identification of the optimal scene graph $\mathcal{G}_{i^*}$ that maximizes $s(\mathcal{G}_i, I)$. 
This optimal graph is identified by simply iterating through all potential scene graphs and selecting the one with the highest similarity.
This approach can be accelerated through the use of spatial partitioning techniques, such as kd-trees, for the preprocessing of node embeddings within the map.

\section{Experiments}

\begin{figure}[t]
    \centering
        \includegraphics[width=0.99\columnwidth]{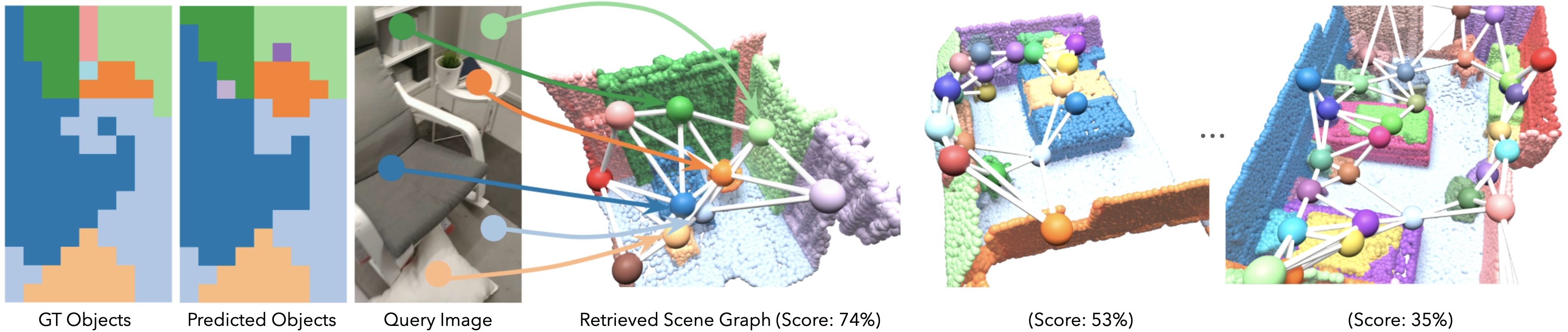}
        \caption{\textbf{Qualitative Result} of object-association-based scene retrieval from the 3RScan dataset~\cite{3rscan}. 
        The two left images show the ground truth (left) and predicted (right) patch-to-node associations of the query image. 
        The right part of the figure illustrates the candidate scene graphs sorted by the image-graph similarity.}
        \label{fig:example}\vspace{-5mm}
\end{figure}

\noindent
\textbf{Baselines.}
No existing methods directly tackle our specific challenge, but several recent advancements provide relevant baselines. LidarCLIP~\cite{hess2024lidarclip}, designed for autonomous driving, transforms LiDAR point clouds into global descriptors using the Single-stride Sparse Transformer~\cite{fan2022sst} and matches them with CLIP image encoder embeddings~\cite{radford2021clip}. 
Though not a perfect fit, it can be adapted for matching point clouds with CLIP embeddings of query images. 
LIP-Loc~\cite{shubodh2024lip} employs a similar approach by converting LiDAR point clouds into 2D range images for direct encoding and matching. 
Both methods were fine-tuned on our dataset for accurate comparison.
We also explored object-retrieval-based baselines using OpenMask3D~\cite{takmaz2023openmask3d} and OpenSeg~\cite{ghiasi2021openseg}, with OpenMask3D assigning CLIP descriptors to 3D object instances from multiple observing images, and OpenSeg extracting pixel-level CLIP features from the query image. 
Despite their original purposes, both can be adapted for localization by matching query image descriptors with object instances.
For comparison with state-of-the-art visual localization methods requiring large image databases, we included CVNet~\cite{lee2022correlation} and AnyLoc~\cite{keetha2023anyloc}. 
These methods offer advanced performance but demand significant storage for image descriptors and exhibit slower inference times.

\noindent
\textbf{Map Generation.} 
The mapping stage is executed offline as a preprocessing step for visual localization approaches, requiring specific mapping operations for each method before proceeding to localization. 
For our proposed method, this entails passing the point cloud, images, metadata, and relationships through the 3D scene graph encoder outlined in Section~\ref{sec:graphencoder}.
For LIP-Loc, this step involves converting point clouds into range images and computing the embeddings of the range images and for LidarCLIP, point clouds are directly encoded into global descriptors.
In OpenMask3D, the CLIP embeddings corresponding to each object are calculated and stored.
For image-based methods like CVNet and AnyLoc, embeddings for all images in the database are precomputed. 

\noindent
\textbf{Metrics.}
To evaluate the accuracy of a method, we focus on the recall of scene selection. This entails analyzing the scenario where, given a query image and corresponding scene pair ($I$, $\mathcal{G}_I$), alongside $N - 1$ alternative scenes from the database, an ordering is established for these scenes according to their computed similarity to $I$ as determined by the tested method. 
The metric Recall@$K$ is employed to ascertain whether the target scene $\mathcal{G}_i$ is ranked among the top $K$ scenes in terms of similarity as identified by the evaluated method.
Additionally, we will report the inference time and storage requirements.


\begin{table}[tb]
    \caption{Retrieval recall on the test set of 3RScan dataset~\cite{3rscan} (\%; target scene ranked within the top 1, 3, and 5 of the retrieved list) and storage requirements (MB) for methods utilizing point clouds ($\mathcal{P}$), images ($\mathcal{I}$), and other modalities (O $ = \{\mathcal{A}, \mathcal{S}, \mathcal{R} \}$) for map representation. $R$ and $R^t$ represent the recall in the static and temporal scenarios respectively.
    Additionally, metrics for single-modal methods (CVNet and AnyLoc) reliant on extensive image datasets are presented. 
    The results are reported for scenarios where the target room is chosen from a subset of 10 and 50 candidate scenes.}
  \label{tab:3drscan_results}
  \centering
  \resizebox{1.0\textwidth}{!}{\begin{tabular}{@{} | l | c c c | ccc | ccc | ccc | ccc | c |@{}}
    \hline
    \multicolumn{1}{|c|}{\multirow{2}{*}{Method}}
     &  \multicolumn{3}{c|}{Map modalities} &  \multicolumn{6}{c|}{10 scenes} & \multicolumn{6}{c|}{50 scenes} & Storage  \\
      & \phantom{--}$\mathcal{P}$\phantom{---} & \phantom{--}$\mathcal{I}$\phantom{--} & O & $R{@1}$ & ${@3}$ & ${@5}$ & $R^t{@1}$ & ${@3}$ & ${@5}$ & $R{@1}$ & ${@3}$ & ${@5} $ & $R^t{@1}$ & ${@3}$ & ${@5}$ & (MB) \\
    \hline
     LidarCLIP~\cite{hess2024lidarclip} & \checkmark & \xmark & \xmark 
        & 16.3 & 41.4 & 60.6 & 16.3 & 39.8 & 61.1 & \phantom{1}4.7 & 11.0 &  16.3 & \phantom{1}4.1 & 10.3 & 15.6 & \phantom{100}\textbf{0.4} \\
     LIP-Loc~\cite{shubodh2024lip} & \checkmark & \xmark & \xmark 
        & 14.0 & 35.8 & 57.9 & 10.9 & 30.0 & 52.7 & \phantom{1}2.0 & \phantom{1}9.1 & 14.2 & \phantom{1}2.3 & \phantom{1}8.6 & 15.2 & \phantom{100}1.0 \\
     OpenMask3D~\cite{takmaz2023openmask3d} & \checkmark & \checkmark & \xmark 
        & -- & -- & -- & 42.3 & 71.5 & 85.8 & -- & -- & -- & 21.1 & 38.1 & 48.0 & \phantom{10}20.1  \\
    \name{} (Ours) & \checkmark & \xmark & \checkmark 
        & \textbf{53.6} & \textbf{81.9}& \textbf{92.8} & 50.5 & 76.8 & 88.4 & \textbf{30.2} & \textbf{50.2} & \textbf{61.2} & 28.2 & 46.2 & 56.4 & \phantom{100}5.4 \\
     \name{} (Ours) & \checkmark & \checkmark & \checkmark
        & -- & -- & -- & \textbf{81.5} & \textbf{93.9} & \textbf{98.0} & -- & -- & -  & \textbf{69.3} & \textbf{78.6} & \textbf{84.4} & \phantom{100}5.4 \\
    \hline
     CVNet~\cite{lee2022correlation} & \xmark & \checkmark & \xmark
        & -- & -- & -- & 79.2 & 91.0 & 95.4 & -- & -- & -- & 66.5 & 77.0 & 81.7 & \phantom{1}239.1 \\
     AnyLoc~\cite{keetha2023anyloc} & \xmark & \checkmark & \xmark 
        & -- & -- & -- & \textbf{87.9} & \textbf{94.7} & \textbf{97.5} & -- & -- & -- & \textbf{80.6} & \textbf{87.4} & \textbf{90.0} & 5720.3 \\
    \hline
  \end{tabular}}
\end{table}

\noindent
\textbf{Experiments on the 3RScan Dataset.}
The 3RScan dataset \cite{3rscan} comprises $1335$ annotated indoor scenes, representing 432 distinct spaces (rooms), with $1178$ scenes (385 rooms) allocated for training and $157$ (47 rooms) designated for validation. 
The training and validation sets include semantically annotated 3D point clouds for each scene, with some captured over extended periods (\eg, several months) showcasing environmental changes. 
Annotations for graphs within the 3RScan dataset are provided in \cite{3dssg}. Due to the absence of such annotations in the test set, it was excluded from our experiments. Thus, we reorganized the original validation set, allocating $34$ scenes (17 rooms) for validation and $123$ scenes (30 rooms) for testing.
For full reproducibility, we will publish this split.


During testing, we examine all 123 scenes of 30 rooms within the test set, selecting query images from each scene.
The next step involves matching this image against $N$ scenes (including the target) to ascertain whether the correct one could be identified by a method.
This procedure is repeated for every image in each room.
In total, all methods are tested on 30462 query images.
%
Furthermore, we evaluate scene selection through two settings $N=50$ and $N=10$. 
The latter setting emulates a scenario where a pre-selection strategy is employed, for example, utilizing a global scene descriptor.
In image-based methods, we use all images from the database and determine the scene based on the retrieved image.


Additionally, the methods are evaluated under both static and temporal conditions. 
In the static scenario, the target scene graph for a given query image originates from the same scan, albeit from a different sequence, to ensure no image overlap. 
Conversely, in the temporal scenario, the scene graph is derived from a sequence captured at a different temporal stage than the query, introducing potential environmental changes.
We do not show results for methods exploiting images in their maps in the static stage. 

The results are reported in Table~\ref{tab:3drscan_results}.
Despite being retrained, LidarCLIP and LIP-Loc display inaccurate results, particularly in scenarios involving the selection of the target room from the entire scene set. 
LIP-Loc barely surpasses random selection. 
Although LidarCLIP exhibits marginally better accuracy, it remains substantially inferior to alternative methods. 
The temporal case further decreases the performance of both methods.
%
OpenMask3D, while achieving better results than LidarCLIP and LIP-Loc, is less accurate than the proposed \name{}. 
SceneGraphLoc, even when excluding the \texttt{image} modality ($\mathcal{I}$), outperforms other cross-modal strategies by a significant margin. 
Incorporating images significantly enhances its performance, positioning it close to that of image-based approaches but with \textit{three orders of magnitude} smaller storage requirements.
Also, the storage of SceneGraphLoc with and without images is the same due to its design of distilling knowledge into fixed-sized embeddings.
An example scene is shown in Fig.~\ref{fig:example}.

\begin{table}[tb]
    \caption{Retrieval recall in the temporal scenario on the test set of ScanNet dataset~\cite{dai2017scannet} (\%; target scene ranked within the top 1, 3, and 5 of the retrieved list) and storage requirements (MB) for methods using point clouds ($\mathcal{P}$), images ($\mathcal{I}$), and other modalities ($\mathcal{S}$, $\mathcal{R}$) for map representation. The modality $\mathcal{A}$ is not available in the scene graphs predicted from \cite{wu2021scenegraphfusion}. 
    The results are reported for scenarios where the target room is chosen from a subset of 10, 50 and the complete set of all (210) scenes.}
  \label{tab:scannet_test_set}
  \centering
  \resizebox{1.0\textwidth}{!}{\begin{tabular}{@{} | l | c c c | ccc | ccc | ccc | c |@{}}
    \hline
    \multicolumn{1}{|c|}{\multirow{2}{*}{Method}}
     &  \multicolumn{3}{c|}{Map modalities} &  \multicolumn{3}{c|}{10 scenes} & \multicolumn{3}{c|}{50 scenes} & \multicolumn{3}{c|}{All scenes} & Storage  \\
      & \phantom{--}$\mathcal{P}$\phantom{---} & \phantom{--}$\mathcal{I}$\phantom{--} & O & $R^t{@1}$ & ${@3}$ & ${@5}$ & $R^t{@1}$ & ${@3}$ & ${@5}$ & $R^t{@1}$ & ${@3}$ & ${@5}$ & (MB) \\
    \hline
     LidarCLIP~\cite{hess2024lidarclip} & \checkmark & \xmark & \xmark 
        & 19.4 & 47.5 & 67.6 & \phantom{1}4.7 & 14.8 & 22.2 & \phantom{1}5.9 & 15.0 & 21.9 & \phantom{100}\textbf{0.7} \\
     LIP-Loc~\cite{shubodh2024lip} & \checkmark & \xmark & \xmark 
        & 10.3 & 27.0 & 43.6 & \phantom{1}1.9 & \phantom{1}6.0 & \phantom{1}8.1 & \phantom{1}1.8 & \phantom{1}3.1 & \phantom{1}4.0 & \phantom{100}1.7\\
     OpenMask3D~\cite{takmaz2023openmask3d} & \checkmark & \checkmark & \xmark 
        & 54.9 & 84.8 & 94.0 & 31.3 & 51.3 & 63.2 & 16.5 & 27.2 & 34.5 & \phantom{10}17.8 \\
    \name{} (Ours) & \checkmark & \xmark & \checkmark 
        & 54.1 & 81.4 & 91.9 & 29.0 & 47.4 & 58.0 & 13.5 & 26.4 & 34.2 & \phantom{100}9.3 \\
     \name{} (Ours) & \checkmark & \checkmark & \checkmark
        & \textbf{78.5} & \textbf{92.7} & \textbf{98.3} & \textbf{61.6} & \textbf{83.2} & \textbf{91.6}  & \textbf{53.4} & \textbf{69.8} & \textbf{78.7} & \phantom{100}9.3 \\
    \hline
     CVNet~\cite{lee2022correlation} & \xmark & \checkmark & \xmark
        & 96.5 & 98.9 & 99.6 & 92.6 & 96.0 & 97.0 & 89.9 & 93.4 & 94.6 & \phantom{1}239.1 \\
     AnyLoc~\cite{keetha2023anyloc} & \xmark & \checkmark & \xmark 
        & \textbf{98.4} & \textbf{99.4} & \textbf{99.8} & \textbf{96.5} & \textbf{98.1} & \textbf{98.6} & \textbf{95.1} & \textbf{96.9} & \textbf{97.4} & 5720.3 \\
    \hline
  \end{tabular}}
\end{table}

\noindent
\textbf{Experiments on the ScanNet Dataset.}
In order to evaluate generalization ability of our methods in real-world applications when scene graph annotations are not available, we conduct further experiments in the ScanNet dataset~\cite{dai2017scannet}.
ScanNet encompasses 1613 monocular sequences of room-scale 3D scenes, offering 3D mesh reconstructions alongside the RGBD frame sequences utilized for the reconstructions. 
Given the absence of scene graph annotations in ScanNet, 
we run the SceneGraphFusion~\cite{wu2021scenegraphfusion} on the RGBD sequences of scans for 3D reconstruction and scene graph prediction with 3D instance segmentation and object relationships (\ie, graph edges) within these graphs.
As the process of scene graph prediction uses the RGBD frames of each scan, we avoid using those RGB images to match to the scene graph predicted of the scan. 
Thus, we only measure recall in the temporal scenario. 
Additionally, unlike 3RScan, the frame rate of RGBD sequences in ScanNet is high, and motion between consecutive frames is small. 
Thus, each database image is selected from every 25 consecutive frames in the sequence, for image-based methods~\cite{lee2022correlation, keetha2023anyloc}. 
For a fair comparison, all the methods only use the same selected images for training and evaluation. 

For training, we use the official training set. 
We divide the official validation set, which includes 312 scenes, into two distinct subsets: the first 100 scenes form our validation set, while the remaining 212 are allocated for testing. 
To ensure full reproducibility, we will make this split publicly available.

The results are in Table~\ref{tab:scannet_test_set} for scenarios selecting the target room from subsets of 10, 50, and the entire set of 210 scenes. 
The performance of LIP-Loc shows a similar pattern to that observed on 3RScan, performing only slightly better than random selection. 
LidarCLIP shows a small improvement in accuracy. 
OpenMask3D attains an accuracy comparable to our proposed method without incorporating the image modality. 
Our proposed SceneGraphLoc, when including the image modality in its map, significantly outperforms all cross-modal approaches. 
Although there remains a gap in accuracy compared with methods that use extensive image collections as maps (such as CVNet and AnyLoc), SceneGraphLoc benefits from a database size \textit{three} orders-of-magnitude smaller, highlighting its efficiency and effectiveness.
We partly attribute this performance gap to the lack of object attributes in the dataset and the inaccurate instance segmentation predicted by \cite{wu2021scenegraphfusion}. More details can be found in the supp.\ mat.

\begin{table}[tb]
  \caption{
  Average time (ms) of obtaining the query image embedding ($t_{e_q}$) and of the retrieval from $10$, $50$, and all scenes from the 3RScan~\cite{3rscan} and ScanNet~\cite{dai2017scannet} datasets.
  }
  \label{tab:runtime}
  \centering
  \begin{tabular}{@{} | l | ccc | cccc | @{}}
    \hline
  \multicolumn{1}{|c|}{\multirow{2}{*}{Method}} & \multicolumn{3}{c|}{3DRScan~\cite{3rscan}} & \multicolumn{4}{c|}{ScanNet~\cite{dai2017scannet}} \\
     & \phantom{1}$t_{e_q}$ & \phantom{1}$t^{10}_\text{retr}$ & \phantom{1}$t^\text{50}_\text{retr}$
              & \phantom{1}$t_{e_q}$ & \phantom{1}$t^{10}_\text{retr}$ & \phantom{1}$t^{50}_\text{retr}$ & \phantom{1}$t^\text{all}_\text{retr}$
              \\
    \hline
     LidarCLIP~\cite{hess2024lidarclip} & \phantom{10}4.1 & \phantom{10}0.1 & \phantom{100}0.3 & \phantom{10}4.9 & \phantom{1}0.1 & \phantom{10}0.2 & \phantom{100}0.6 \\
     LIP-Loc~\cite{shubodh2024lip} & \phantom{10}2.7 & \phantom{10}0.1 & \phantom{100}0.2 & \phantom{10}4.1 & \phantom{1}0.1& \phantom{10}0.2 & \phantom{100}0.5 \\
     OpenMask3D~\cite{takmaz2023openmask3d} & \phantom{1}41.5 & \phantom{10}4.8 & \phantom{100}7.4 &\phantom{1}20.1  & 55.4 & \phantom{10}1.1 & \phantom{100}4.5  \\
     \name{} (Ours) & \phantom{1}28.0 & \phantom{10}0.3 & \phantom{100}1.5 & \phantom{1}16.6 & \phantom{1}1.3 & \phantom{10}3.7 & \phantom{10}17.0 \\ 
    \hline
     CVNet~\cite{lee2022correlation} & \phantom{1}14.3 & \phantom{10}9.0 & \phantom{10}60.0  
        & \phantom{1}54.0 & 10.6 & \phantom{1}74.1 & \phantom{1}311.3 \\
     AnyLoc~\cite{keetha2023anyloc} & 658.4 & 354.6 & 1826.4 & 243.0 &  68.2& 329.0& 1451.1\\
    \hline
  \end{tabular}
\end{table}
\noindent
\textbf{Processing time} measured in milliseconds for various methods applied to the 3DRScan and ScanNet datasets, are detailed in Table~\ref{tab:runtime}. 
The computation time required to generate an embedding for the query image ($t_{e_q}$) is notably small across all methods, typically not exceeding a few tens of milliseconds, with the exception of AnyLoc, which runs for nearly a second.
The retrieval phase for cross-modal approaches is generally limited to a few tens of milliseconds. 
However, methods such as CVNet and AnyLoc exhibit slower performance, due to searching through extensive image collections. 
When tasked with selecting from a large number of images, the processing times of these methods can extend into the range of several hundred milliseconds or even reach upwards of a second.

\begin{table}[tb]
  \centering
  \caption{Ablation study performed on the val.\ split of 3RScan~\cite{3rscan}, analysing map modalities ($\mathcal{P}$ -- point cloud, $\mathcal{I}$ -- image, $\mathcal{A}$ -- attributes, $\mathcal{S}$ -- structure, $\mathcal{R}$ -- relationships) and the method (Dino v2 or GCVit) to obtain the image embeddings. } \label{table: albation_modalities}
  \label{tab:headings}
  \begin{tabular}{|@{} ccccc | ccc | ccc | ccc | ccc |@{}}
    \hline
     \multicolumn{5}{|c|}{\multirow{1}{*}{Map modalities}}            & \multicolumn{6}{c|}{Dino v2~\cite{oquab2023dinov2}}                             & \multicolumn{6}{c|}{GCVit~\cite{hatamizadeh2023global}} \\
      \phantom{-}$\mathcal{P}$\phantom{-} & \phantom{-}$\mathcal{I}$\phantom{-} & \phantom{-}$\mathcal{A}$\phantom{-} & \phantom{-}$\mathcal{S}$\phantom{-} & \phantom{-}$\mathcal{R}$\phantom{-} & $R{@1}$ & ${@3}$ & ${@5}$ & $R^t{@1}$ & ${@3}$ & ${@5}$ & $R{@1}$ & ${@3}$ & ${@5}$ & $R^t{@1}$ & ${@3}$ & ${@5}$ \\
    \hline
     \checkmark & & & & & 45.2  & 81.9 & 93.7 & 43.9 & 79.5 & 91.4 
        & 24.6  & 56.0 & 76.9  & 23.2 & 54.7  & 77.3\\
     \checkmark & & \checkmark & & & 56.3 & 85.6 & 95.0 & 54.8 & 84.0 & 95.0
        & 44.2  & \textbf{76.9} & 91.2  & 43.4 & 75.1  & 89.4 \\
     \checkmark & & \checkmark & \checkmark & & 58.4 & \textbf{87.3} & \textbf{95.9} & 56.5 & 85.7 & 93.6
        & 43.3  & 75.8 & \textbf{91.3}  & 41.5 & 72.8  & 89.3\\
     \checkmark & & \checkmark & \checkmark & \checkmark & \textbf{63.7} & 86.8 & 95.8 & 62.7 & 87.4 & 96.3
        & \textbf{45.3}  & 75.5 & 90.5  & 46.6 & 76.2  & 90.2\\
      & \checkmark & & & & -- & -- & -- &80.2 & 96.0  & 99.0 
        & -- & -- & -- & 69.4 & 87.4 &93.7 \\
     \checkmark & \checkmark & & & & -- & -- & -- & 84.7 & 97.5 & \textbf{99.6}
        & -- & -- & -- & \textbf{73.2} & \textbf{89.7}  & \textbf{95.9} \\
     \checkmark & \checkmark & \checkmark & \checkmark & \checkmark  & -- & -- & -- & \textbf{88.5} & \textbf{97.7} & \textbf{99.6}
        & -- & -- & -- & 72.1 & 88.8  & 95.7 \\
    \hline
  \end{tabular}
\end{table}

\noindent
\textbf{Ablation Studies.}
In order to understand the impact of the integration of different modalities within the pipeline, we provide an ablation study on the localization performance with multiple combinations of modalities. 
We use the validation split of the 3DRScan dataset. 
Additional ablation studies can be found in supplementary material.

Table~\ref{table: albation_modalities} displays the coarse localization performance of our method under the incorporation of distinct modalities ($\mathcal{P}$, $\mathcal{I}$, $\mathcal{A}$, $\mathcal{S}$, $\mathcal{R}$) within the pipeline. 
Additionally, the table reports results exploiting various backbones (Dino V2~\cite{oquab2023dinov2} and GCVit~\cite{hatamizadeh2023global}) for the extraction of image features.
Similarly to the main experiments, we only show results on the temporal set when the map contains images. 
Employing Dino V2 for encoding both the query image and images within the map significantly enhances accuracy over using GCVit. 
The proposed method significantly outperforms both LidarCLIP and LIP-Loc (their results are in Table~\ref{tab:3drscan_results}) even when using only point clouds in the map.
Each additional modality contributes to the final accuracy, demonstrating that each plays a crucial role in enhancing the final performance when integrated into the pipeline.


%

\section{Conclusion}

In conclusion, we introduce \name{}, a novel method for solving the novel problem of localizing an input image within a 3D scene graph-based multi-modal reference map. 
This approach outperforms existing cross-modal methods by a large margin. 
It achieves comparable accuracy to state-of-the-art image-based techniques with significantly lower storage requirements and faster processing speeds. 
Our experiments across the 3RScan and ScanNet datasets demonstrate the effectiveness of \name{}, with the best performance achieved when integrating all proposed modalities. 
We believe that \name{} is a step towards lightweight and efficient localization.
The code will be made public.

\title{Supplementary Material \\
SceneGraphLoc}
\author{Yang Miao$^{1}$, Francis Engelmann$^{1,2}$,  Olga Vysotska$^{1}$, Federico Tombari$^{2,3}$
Marc Pollefeys$^{1,4}$, Dániel Béla Baráth$^{1}$}
\institute{}
\maketitle
\begin{abstract}
    In the supplementary material, we provide additional information of \name{}:
    \begin{enumerate}
        \item Qualitative results to further understand the performance of \name{} (Section~\ref{sec: qualitative_results}).
        \item Ablation studies and analysis of experiment results (Section~\ref{sec: ablation_study}).
        \item Details on implementation (Section~\ref{sec: impl_details}). 
    \end{enumerate}
\end{abstract}
\section{Qualitative Results}
\label{sec: qualitative_results}

In this section, we provide additional qualitative results of successful and failed cases of room retrieval with $R^t@1$ in different scenarios. 
\par
\begin{figure}[ht]
    \centering
    \begin{subfigure}[b]{0.99\textwidth}
        \centering
        \includegraphics[width=\textwidth]{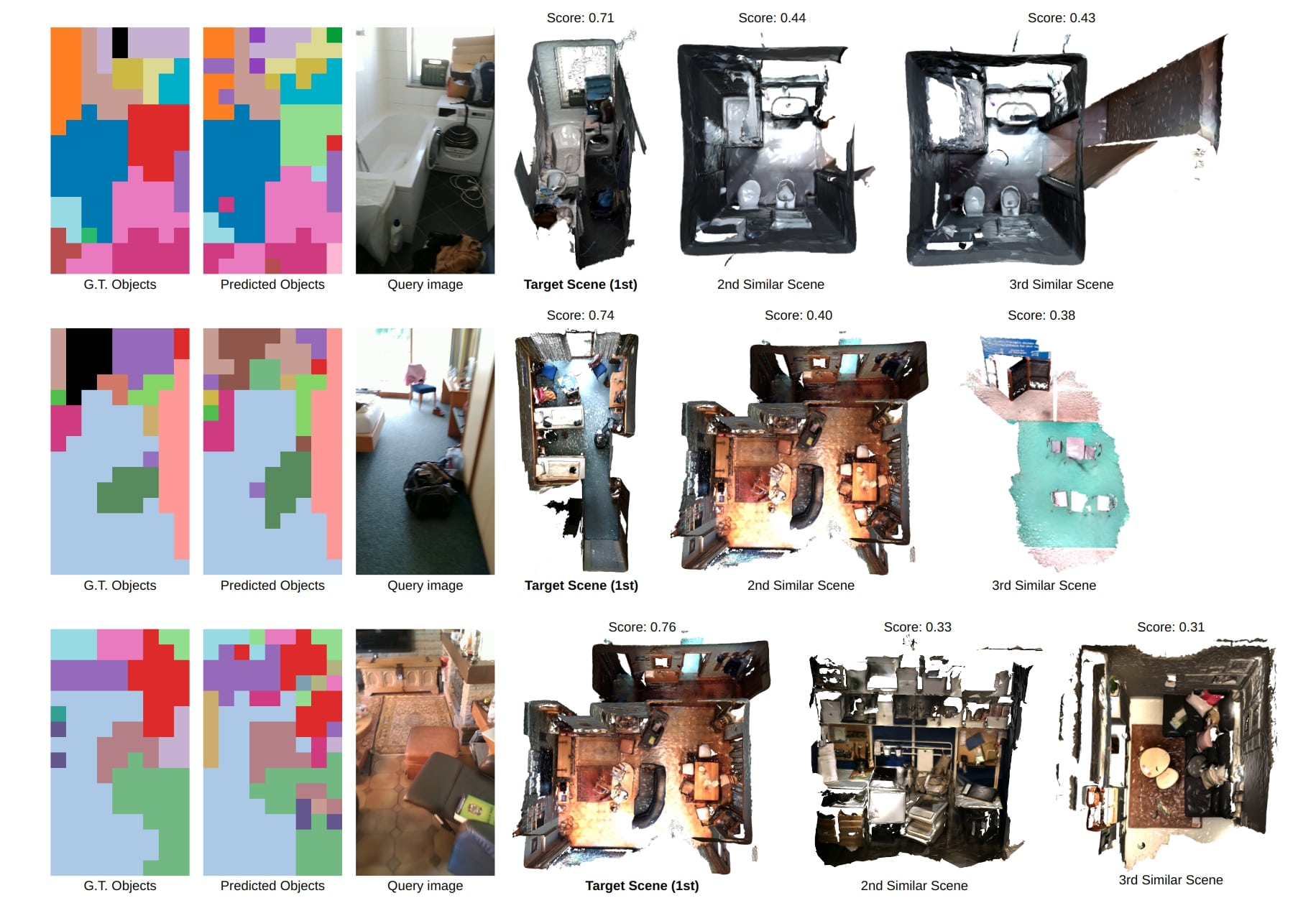} 
        \caption{Successful cases of top-1 scene retrieval}
        \label{fig:qualitative_positive}
    \end{subfigure}
    
    \begin{subfigure}[b]{0.99\textwidth}
        \centering
        \includegraphics[width=\textwidth]{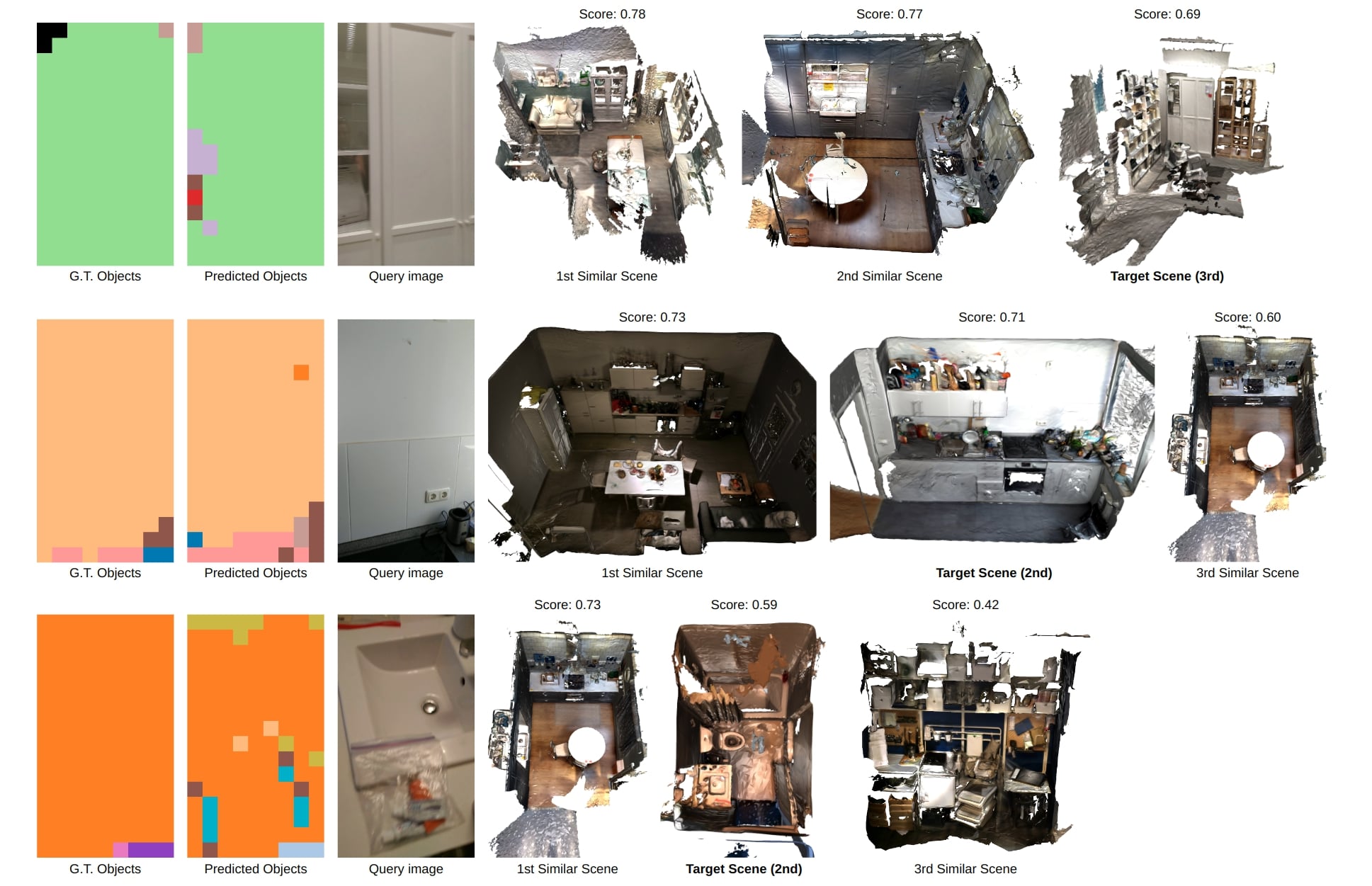} 
        \caption{Failures for top-1 scene retrieval}
        \label{fig:qualitative_negative}
    \end{subfigure}

    \caption{\textbf{Successful} and \textbf{failed} cases for scene retrieval with $R^t@1$. 
    On the left are the G.T.\ and predicted objects of query image within the target scene graph. On the right are the top-3 retrieved scenes with their image-scene similarity scores.}
    \label{fig:qualitative_results}
\end{figure}

In Fig.~\ref{fig:qualitative_positive}, we show successful cases with the target scene ranked as No.1 in the pool of 10 candidate scenes. 
From the left images in Fig.~\ref{fig:qualitative_positive}, we can see that the majority of the patches is correctly assigned to corresponding objects given a pool of objects within the target scene graph. 
Furthermore, the similarity score gap of the target scene is significantly larger than the second most similar scene, showing the effectiveness of the proposed similarity score in distinguishing the target scene and other scenes.
\par 
In Fig.~\ref{fig:qualitative_negative}, we show failure cases with the target scene not ranked as No.1 in the pool of candidate scenes. 
Compared to Fig.~\ref{fig:qualitative_positive}, we can see that the query images in Fig.~\ref{fig:qualitative_negative} have a limited field of view and a limited diversity of observed distinct objects.
The intuition is that the localization performance of the query image is related to the diversity of objects observed in the image: the more diverse and distinctive objects are in the query image, the easier it is for the query image can be correctly matched to the target scene, as shown in Fig.~\ref{fig:qualitative_positive}. 
Conversely, if the query image is dominated by non-unique objects (i.e., wall), then it can be difficult to retrieve the target scene graph, as shown in Fig.~\ref{fig:qualitative_negative}. 
The dependence on the object number will be shown in the next section.
This tendency can be exploited in practice as confidence in the predicted results, assigning high confidence if many objects are seen from the query image. 

\clearpage
\section{Additional Ablation Study}
\label{sec: ablation_study}

\subsection{Image Modality Embedding $\mathcal{I}$}
\begin{table}[t]
    \centering
    \caption{Ablation study on the methods generating image embeddings for the map.}
    \label{table:ablation_study_image_modality}
    \begin{subtable}{.39\linewidth} 
        \centering
        \begin{tabular}{rccc}
            \hline
            $K_{view}$ & \phantom{-}$R^t@1$\phantom{-} & \phantom{-}$@3$\phantom{-} & \phantom{-}$@5$\phantom{-} \\
            \hline
            1 & 83.9 & 96.6 & 99.4 \\
            3 & 84.2 & 96.7 & 99.5 \\
            5 & 85.4 & 97.4 & 99.5 \\
            7 & 86.7  & 97.0 & 99.4\\
            10 & \textbf{88.5} & 97.7 & 99.6 \\
            15 & 86.3 & \textbf{97.8} & 99.6 \\
            20 & 86.8 & \textbf{97.8} & \textbf{99.7} \\
            \hline
        \end{tabular}
        \caption{The recall values w.r.t.\ the number ($K_{view}$) of views used to create an image embedding for a particular object.}
        \label{subtab:K_variation}
    \end{subtable}%
    \hfill
    \begin{subtable}{.59\linewidth} 
        \centering
        
        \begin{tabular}{cccc | ccc}
            \hline
            \multicolumn{4}{c|}{Configuration} & \multirow{2}{*}{$R^t@1$} & \multirow{2}{*}{$@3$} & \multirow{2}{*}{$@5$} \\
            TE & PE & Max & Mean & & & \\ 
            \hline
            \xmark & \xmark &  \xmark & \checkmark & 85.5 & 96.8 & 99.4 \\
            \xmark & \xmark & \checkmark & \xmark & 86.0 & 97.1 & 99.4\\
            \checkmark & \xmark & \checkmark & \xmark & 86.6 & 97.2 & 99.4 \\
            \checkmark & \checkmark &  \checkmark & \xmark & \textbf{88.5} & \textbf{97.7} & \textbf{99.6} \\
            \hline
        \end{tabular}
        \caption{Multi-view image fusion. 
        "Max" and "Mean" indicate max- and average-pooling over the $K_{view}$ views, respectively.
        "TE" indicates using the transformer encoder.
        "PE" means using camera poses for positional encoding in "TE".}
        \label{subtab:multiview_fusion}
    \end{subtable}\vspace{-2mm}
\end{table}
\name{} integrates multi-view image features for object embedding of image modality $\mathcal{I}$, as shown in Fig.3 in the main paper.
In Table~\ref{table:ablation_study_image_modality}, we explore the impact of the number ($K_{view}$) of views considered when creating the multi-view embedding and the employed image fusion methods on the localization performance. Table~\ref{subtab:K_variation} shows that by using more views for modality $\mathcal{I}$, the localization performance improves, and this trend stoping when $K_{view}$ reaches $15$ and $20$. 
Furthermore, Table~\ref{subtab:multiview_fusion} shows that the localization performance benefits from the transformer encoder with positional encoding followed by max-pooling. The intuition behind positional encoding with image poses is to integrate spatial context with the multi-view visual information for more-informed visual embedding of objects within the scene graph. 

\subsection{Correlation between variables and the recall performance}
\begin{figure}[t]
    \centering
        \includegraphics[width=0.99\columnwidth]{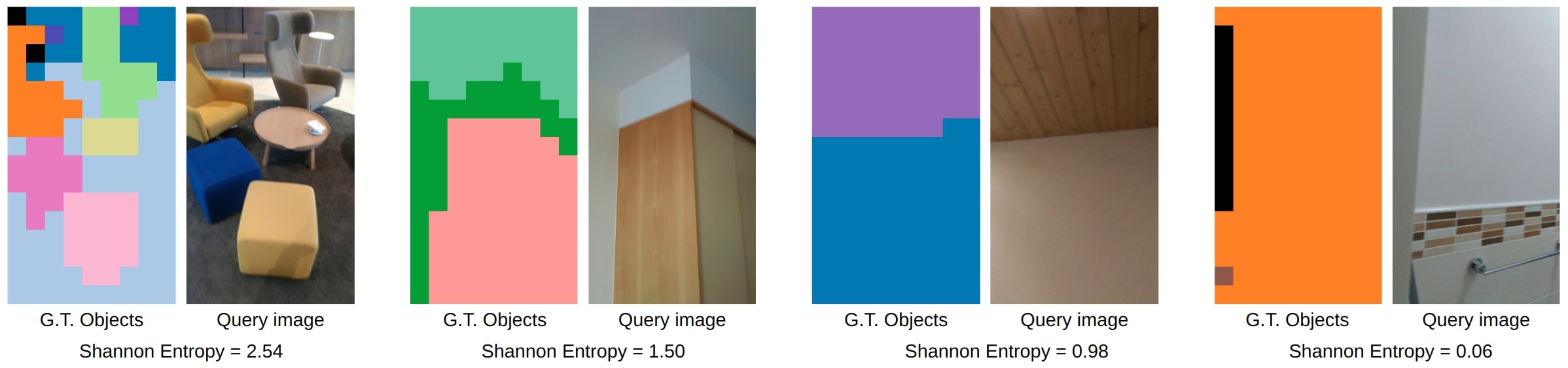}
        \caption{Shannon entropy $\mathcal{H}_I$, denoting the diversity of objects observed in the query image.}
        \label{fig:shannon_entropy}\vspace{-3mm}
\end{figure}

\begin{table}[tb]
  \caption{Statistics Analysis on the val.\ split of 3RScan~\cite{3rscan}, analysing the correlation between multiple factors ($|\mathcal{V}^t_0|$,  $\mathcal{H}_I$ and $s_I$) and the performance of coarse localization ($R^t{@1}$ abbreviated as $R^t_1$ and $Acc_q^t$) under multiple modalities. }
   \label{table: albation_statistics}
  \label{tab:headings}
  \centering
  \begin{tabular}{|@{} ccccc | c c c c | c c c |@{}}
    \hline
     \multicolumn{5}{|c|}{\multirow{1}{*}{Map modalities}} 
        &  \phantom{-}\multirow{2}{*}{$R^t{@1}$}  & \phantom{-}\multirow{2}{*}{$ \rho_{(|\mathcal{V}^t_0|, R^t_1)}$} & \phantom{-}\multirow{2}{*}{$ \rho_{(\mathcal{H}_I, R^t_1)}$} & \phantom{-}\multirow{2}{*}{$\rho_{(s_I, R^t_1)}$ }
        & \phantom{-}\multirow{2}{*}{$Acc_q^t$} & \phantom{-}\multirow{2}{*}{ $ \rho_{(|\mathcal{V}^t_0|,Acc^t_q )}$} & \phantom{-}\multirow{2}{*}{$ \rho_{(\mathcal{H}_I, Acc^t_q )}$}  \\
      \phantom{-}$\mathcal{P}$\phantom{-} & \phantom{-}$\mathcal{I}$\phantom{-} & \phantom{-}$\mathcal{A}$\phantom{-} & \phantom{-}$\mathcal{S}$\phantom{-} & \phantom{-}$\mathcal{R}$\phantom{-} & & & & & & & \\
    \hline
     \checkmark & & & & 
        & 43.9 & 0.20 & 0.16 & 0.02 & 49.2 & -0.10 & 0.03 \\
     \checkmark & & \checkmark & & 
        & 54.8 & 0.21 & 0.19 & 0.07 & 53.8 & -0.06 & 0.05\\
     \checkmark & & \checkmark & \checkmark & 
        & 56.5 & 0.29 & 0.20 & 0.11 & 55.9 & -0.04 & 0.03\\
     \checkmark & & \checkmark & \checkmark & \checkmark 
        & 62.7 & 0.38 & 0.22 & 0.19 & 54.8 & -0.07 & 0.06\\
      & \checkmark & & & & 80.2 & 0.15 & 0.06 & 0.19 & 55.6 & -0.06 & -0.07 \\
     \checkmark & \checkmark & & & & 84.7 & 0.21 & 0.19 & 0.20 & 61.1 & -0.06 & -0.01  \\
     \checkmark & \checkmark & \checkmark & \checkmark & \checkmark  
        & 88.5 & 0.28 & 0.15 & 0.17 & 64.2 & -0.07 & -0.03 \\
    \hline
  \end{tabular}
\end{table}
In Table~\ref{table: albation_statistics}, we report the correlation between multiple factors and the localization performance $R^t{@1}$ and $Acc_q^t$ under multiple settings of modalities.
The following notations are defined:
\begin{itemize}
    \item Scalar $|\mathcal{V}^t_0|$ represents the number of object nodes within the target scene  graph with potential temporal changes $\mathcal{G}^t_I$. 
    \item Scalar $\mathcal{H}_I$ represents the Shannon entropy of object information observed in image patches $q \in \mathcal{Q}_I$, defined in Eq.~\ref{eq:obj_entropy_query_image}. 
    \item Scalar $s_I = s(\mathcal{G}^t_I, I)$ represents the similarity score between $\mathcal{G}^t_I$ and the query image. 
    \item Scalar $Acc_q^t$ represents the percentage of image patches $q_I$ that are correctly assigned to the objects in the scene graph given Eq. 3 in the main paper.
    \item Scalar $\rho(a, b) \in [-1, 1]$ represents the Pearson Correlation coefficient between two variables $a$ and $b$. Parameter $\rho>0$ represents positive correlation while $\rho<0$ means negative correlation.
\end{itemize}
\begin{equation} \label{eq:obj_entropy_query_image}
    \begin{split}
        \mathcal{H}_I &= -\sum_{o \in \mathcal{O}^{gt}_I} p_I(o)\text{log}p_I(o), \\  
        p_I(o) &= \frac{|\{q_I \vert q_I \in \mathcal{Q}_I, o_I(q_I) = o \}|}{|\mathcal{Q}_I|}, 
    \end{split}
\end{equation}
$\mathcal{O}^{gt}_I$ is the ground truth set of objects observed in query image $I$ and $p_I(o)$ is the frequency of patches observing the object $o$. Scalar $\mathcal{H}_I$ denotes the diversity of objects observed in $I$, as illustrated in Fig.~\ref{fig:shannon_entropy}.
\par
From the table, we can see that: 
\begin{itemize}
    \item Values $ \rho_{(|\mathcal{V}^t_0|, R^t_1 )}$ and $ \rho_{(|\mathcal{H}_I|, R^t_1 )}$ are greater than $0$ by a not negligible amount, denoting positive correlation between $|\mathcal{V}^t_0|$ and $R^t_1$, and the positive correlation between $\mathcal{H}_I$ and $R^t_1$. The intuition is that the more objects observed in the query image and located in the target 3D scene graph, the easier the query image can be localized. This correlation agrees with the qualitative results in Section~\ref{sec: qualitative_results}.
    \item Noticeably, with integration of modalities $\{\mathcal{S}, \mathcal{R}\}$, the correlation $\rho_{(|\mathcal{V}^t_0|, R^t_1)}$, $ \rho_{(\mathcal{H}_I, R^t_1)}$ increases. The intuition is that by incorporating $\{\mathcal{S}, \mathcal{R}\}$, the proposed modules learn to leverage scene-context information, e.g., the relationship between objects, for object embedding and coarse localization. Thus, the localization accuracy $R^t_1$ benefits from more context information (larger $\rho_{(|\mathcal{V}^t_0|, R^t_1)}$ and $ \rho_{(\mathcal{H}_I, R^t_1)}$). 
    \item For patch-object association accuracy $Acc_q^t$, all modalities except $\mathcal{R}$ have contributions to improving $Acc_q^t$. On the other hand, there is a slightly negative correlation between $|{V}^t_0|$ and $Acc_q^t$, denoting that the more diversity of the objects in the scene graph, the slightly harder for the image patches to be correctly assigned to certain objects. Noticeably, with integration of image modality $\mathcal{I}$, the correlation $ \rho_{(\mathcal{H}_I, Acc^t_q )}$ turns from slightly positive to slightly negative, denoting that with object embedding of $\mathcal{I}$, the diversity of objects observed in the query image affects the patch-object matching accuracy. 
\end{itemize}

\subsection{The Impact of 3D Instance Segmentation Accuracy}
\begin{figure}[ht]
    \centering
    \begin{subfigure}[b]{0.45\textwidth}
        \centering
        \includegraphics[width=\textwidth]{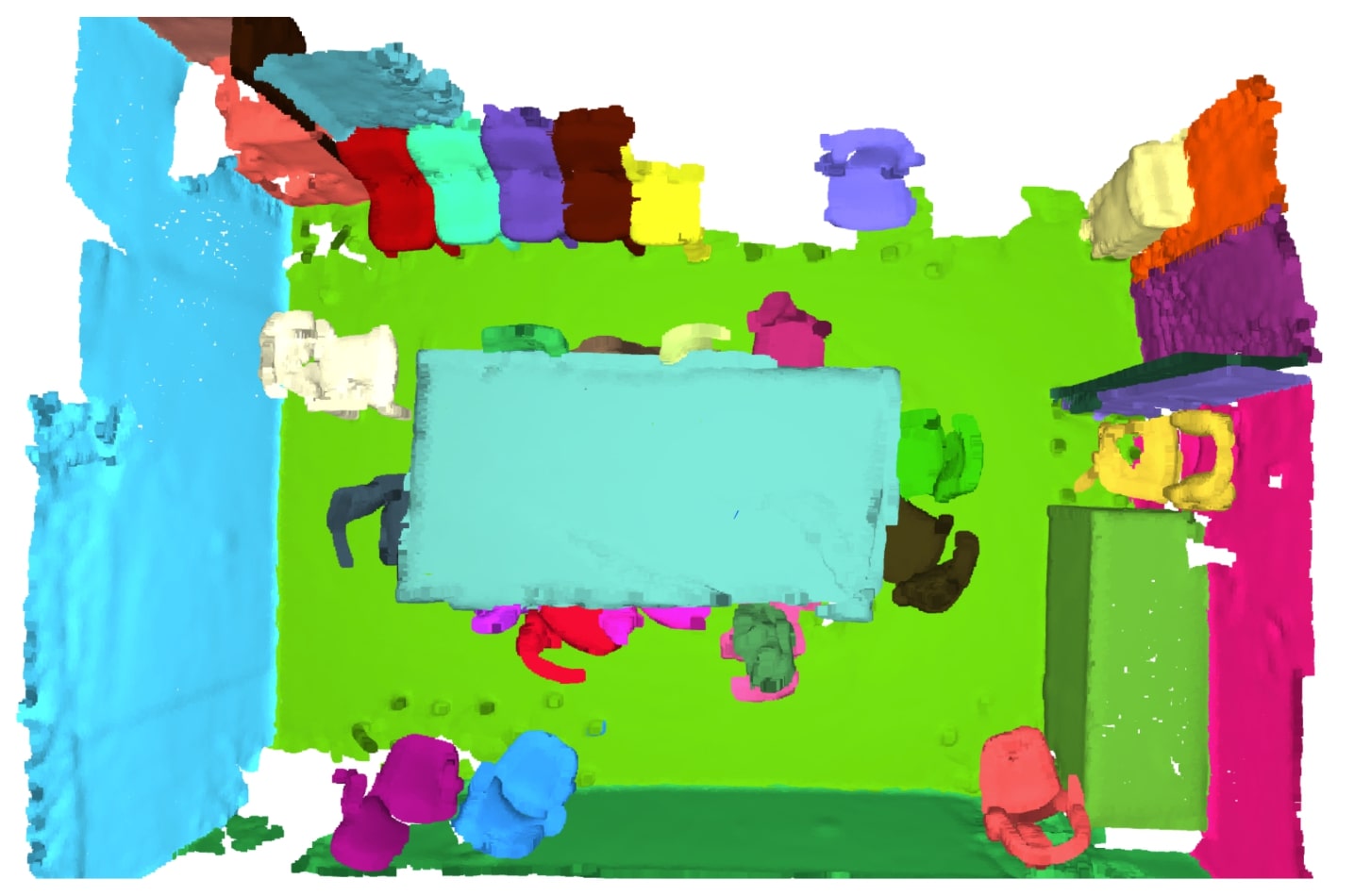} 
        \caption{G.T. Instance Segmentation.}
        \label{fig:scannet_gt_seg}
    \end{subfigure}
    \hfill 
    \begin{subfigure}[b]{0.45\textwidth}
        \centering
        \includegraphics[width=\textwidth]{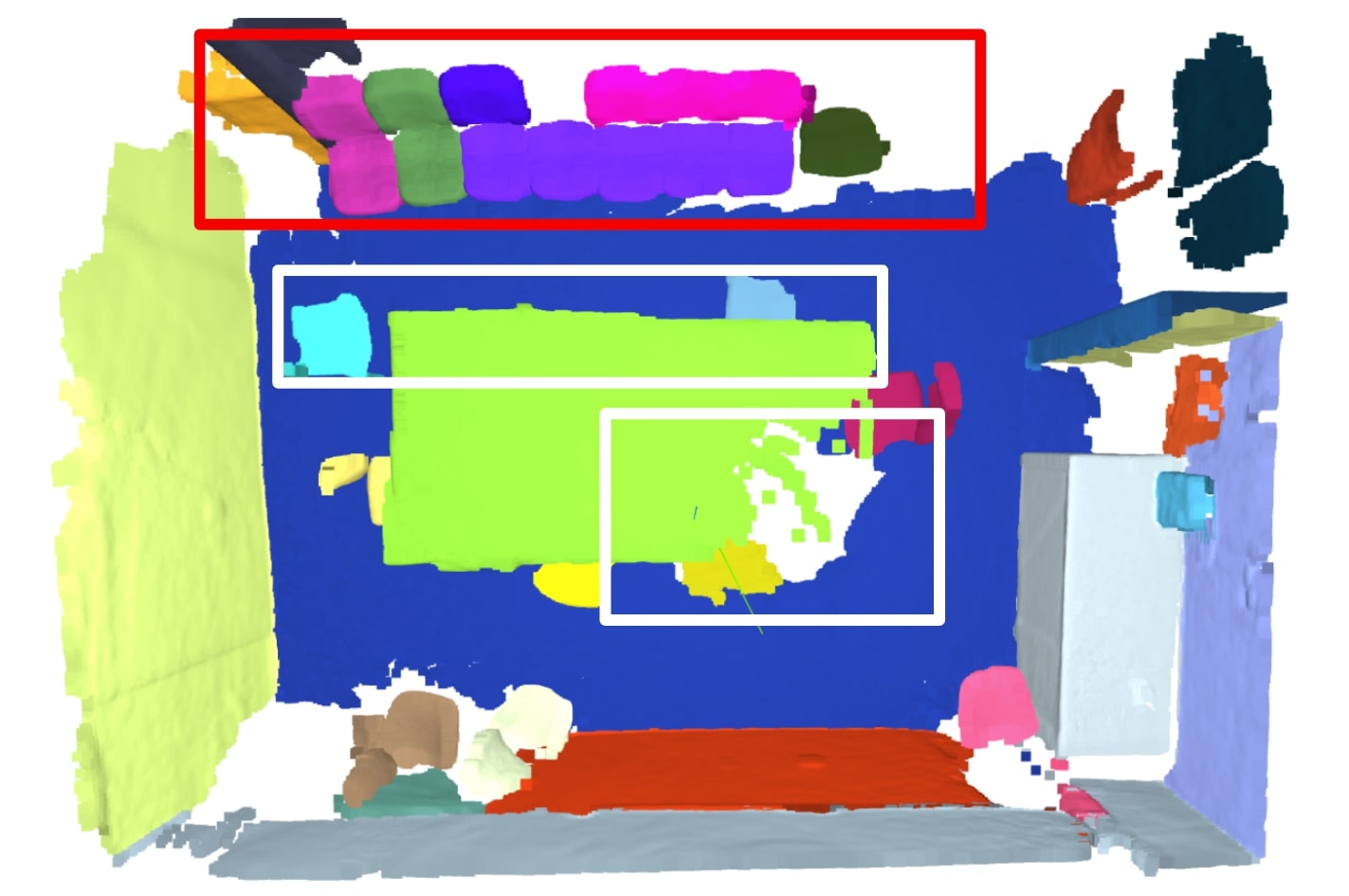} 
        \caption{Predicted Instance Segmentation~\cite{wu2021scenegraphfusion}.}
        \label{fig:confugsion_matrices_sub2}
    \end{subfigure}

    \caption{Comparison of G.T. and predicted instance segmentation in ScanNet dataset~\cite{dai2017scannet}. 
    The left image shows that SceneGraphFusion~\cite{wu2021scenegraphfusion} applied in the Section 4 in the main paper can output inaccurate instance segmentation (red box) and under-reconstruction (white boxes) results. }
    \label{fig:scannet_instance_seg}
\end{figure}

\begin{table}[tb]
  \centering
  \caption{Ablation study performed on the val.\ split of ScanNet~\cite{dai2017scannet} with \name{} with ground truth 3D instance segmentation and predicted instance segmentation from~\cite{wu2021scenegraphfusion}. } \label{table: albation_gt_seg}
  \label{tab:headings}
  \begin{tabular}{|@{} cc | ccc | ccc |@{}}
    \hline
     \multicolumn{2}{|c|}{\multirow{1}{*}{Map modalities}}            & \multicolumn{3}{c|}{G.T. Seg}                             & \multicolumn{3}{c|}{Predicted Seg~\cite{wu2021scenegraphfusion}} \\
      \phantom{----}$\mathcal{P}$\phantom{-} & $\mathcal{I}$\phantom{-} & \phantom{-}$R^t{@1}$\phantom{-} & \phantom{-}${@3}$\phantom{-} & \phantom{-}${@5}$\phantom{-} & \phantom{-}$R^t{@1}$\phantom{-} & \phantom{-}${@3}$\phantom{-} & \phantom{-}${@5}$\phantom{-} \\
    \hline
      \phantom{----}\checkmark & & 62.0 & 86.9 & 94.5 & 53.1 & 85.1 & 93.4 \\
      \phantom{----}\checkmark &\checkmark & 79.3 & 96.3 & 99.4 & 68.7 & 94.9 & 98.8\\
    \hline
  \end{tabular}
\end{table}
In the main paper, the experiments on ScanNet\cite{dai2017scannet} with predicted scene graph with \cite{wu2021scenegraphfusion} shows that there is a performance gap between \name{} and the image-retrieval-based methods. 
One potential reason for the gap is the inaccurate instance segmentation from \cite{wu2021scenegraphfusion}, as illustrated in Fig.~\ref{fig:scannet_instance_seg}, as the object embedding of modalities $\mathcal{P}$ and $\mathcal{I}$ requires a 3D model of each object node within the scene graph. 
In order to understand the impact of 3D instance segmentation accuracy in the performance, we compare the performance of \name{} with predicted and ground truth 3D instance segmentation under the modalities of object embedding ($\mathcal{P}$ and $\mathcal{I}$). From Table~\ref{table: albation_gt_seg} we can see that by using ground truth instance segmentation, the performance of \name{} improves by a large margin, implying that the performance in Table 2 in the main paper can be potentially improved by applying more accurate 3D instance segmentation methods when creating the reference map of the environment.

\subsection{Confusion Matrix}
\begin{figure}[t]
    \centering
        \includegraphics[width=0.99\columnwidth]{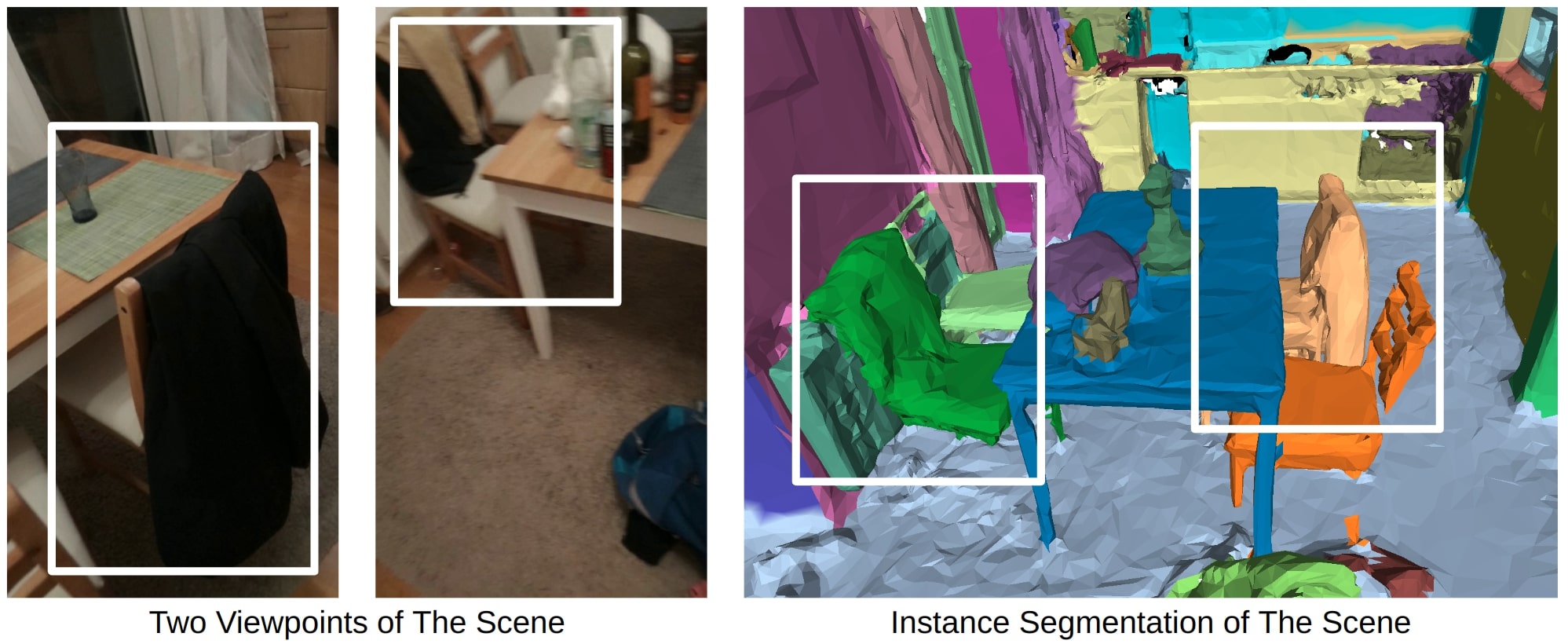}
        \caption{The two left image show that the clothes hangs on the chair back and the right image shows the under-segmentation of the clothes and the chairs.}
        \label{fig:clothes_on_chair}\vspace{-3mm}
\end{figure}

\begin{figure}[ht]
    \centering
    \begin{subfigure}[b]{0.45\textwidth}
        \centering
        \includegraphics[width=\textwidth]{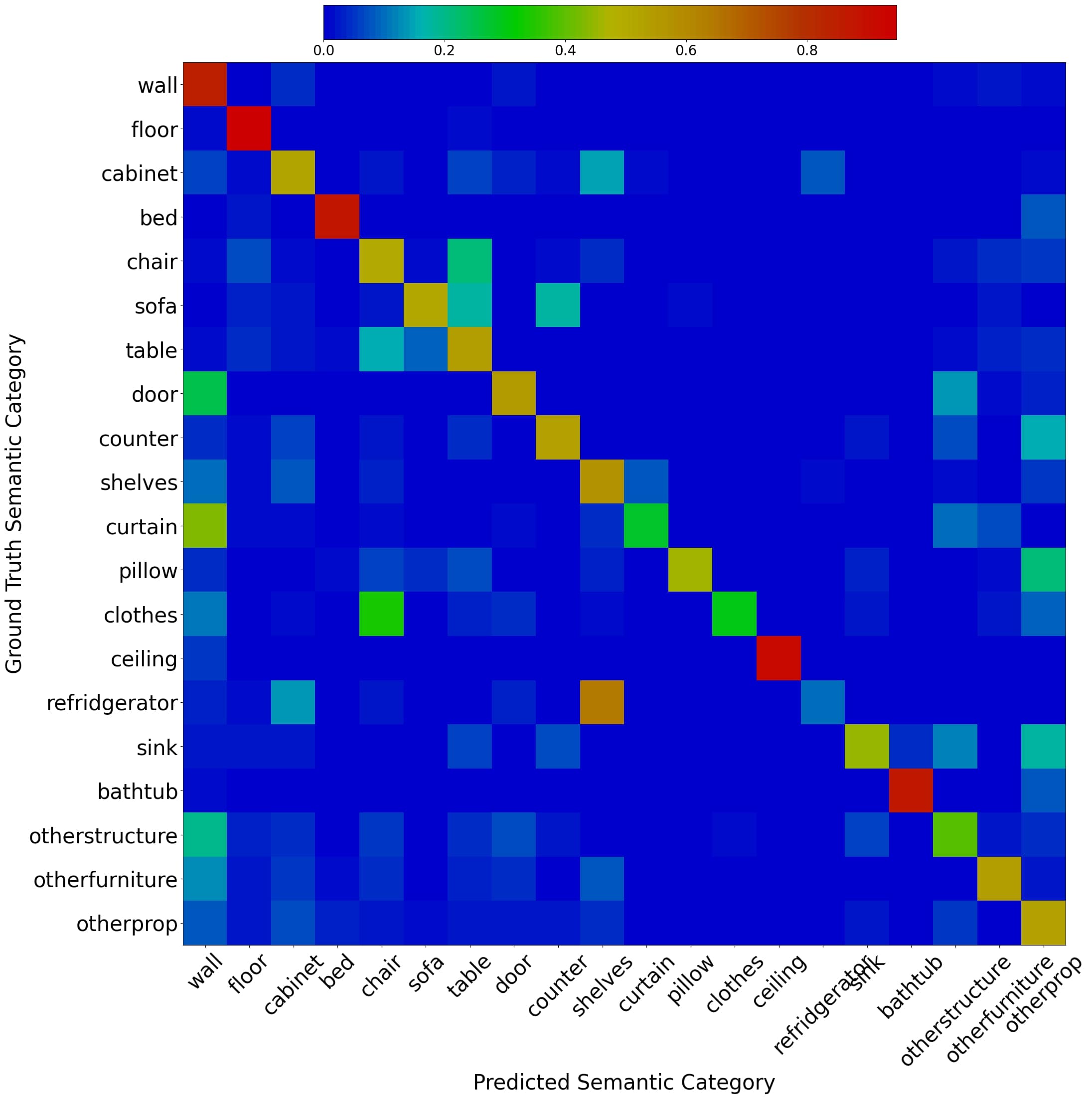} 
        \caption{$\mathcal{P}$}
        \label{fig:confugsion_matrices_sub1}
    \end{subfigure}
    \hfill 
    \begin{subfigure}[b]{0.45\textwidth}
        \centering
        \includegraphics[width=\textwidth]{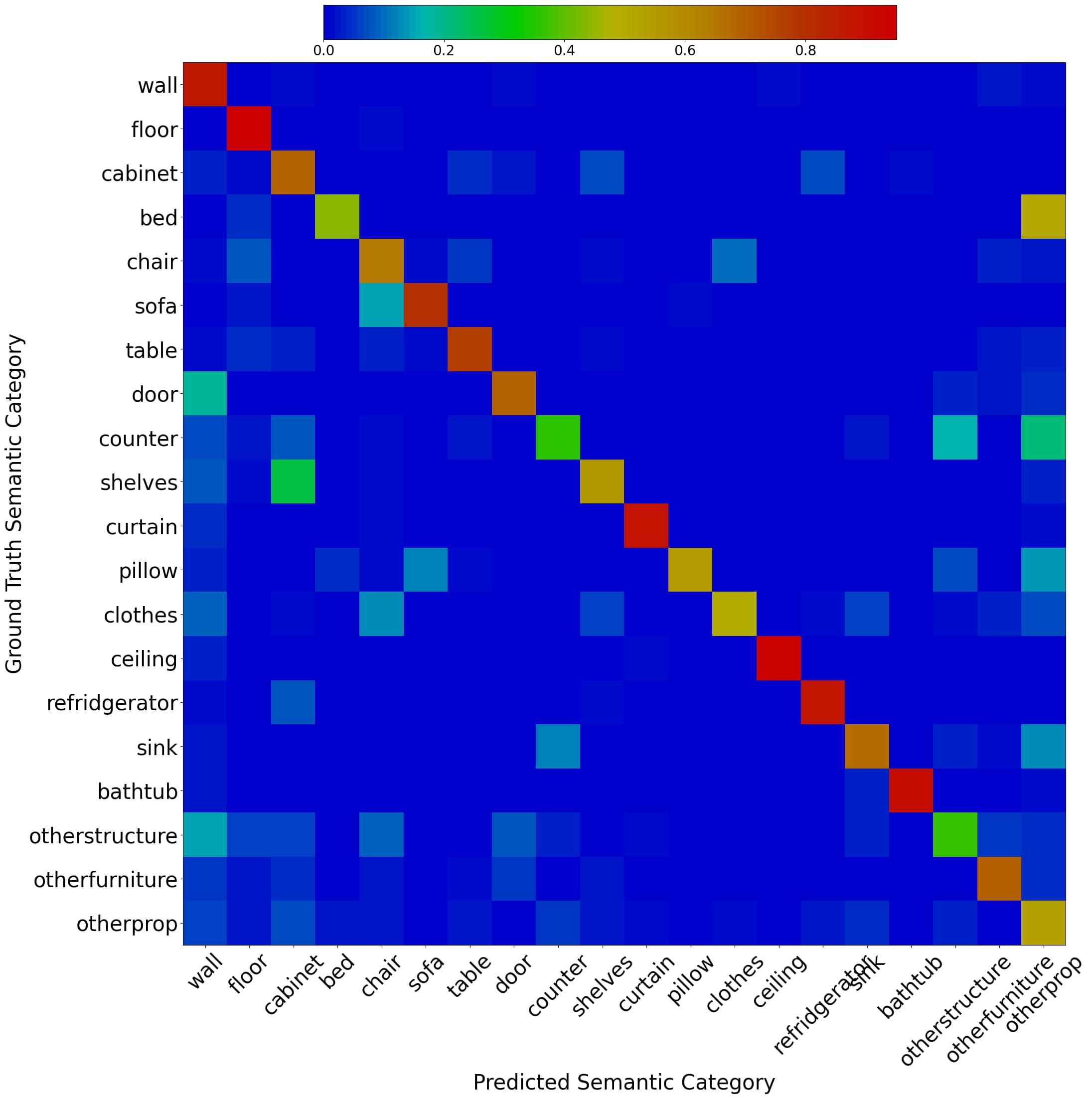} 
        \caption{$\mathcal{P} + \mathcal{A}$}
        \label{fig:confugsion_matrices_sub2}
    \end{subfigure}
    
    \begin{subfigure}[b]{0.45\textwidth}
        \centering
        \includegraphics[width=\textwidth]{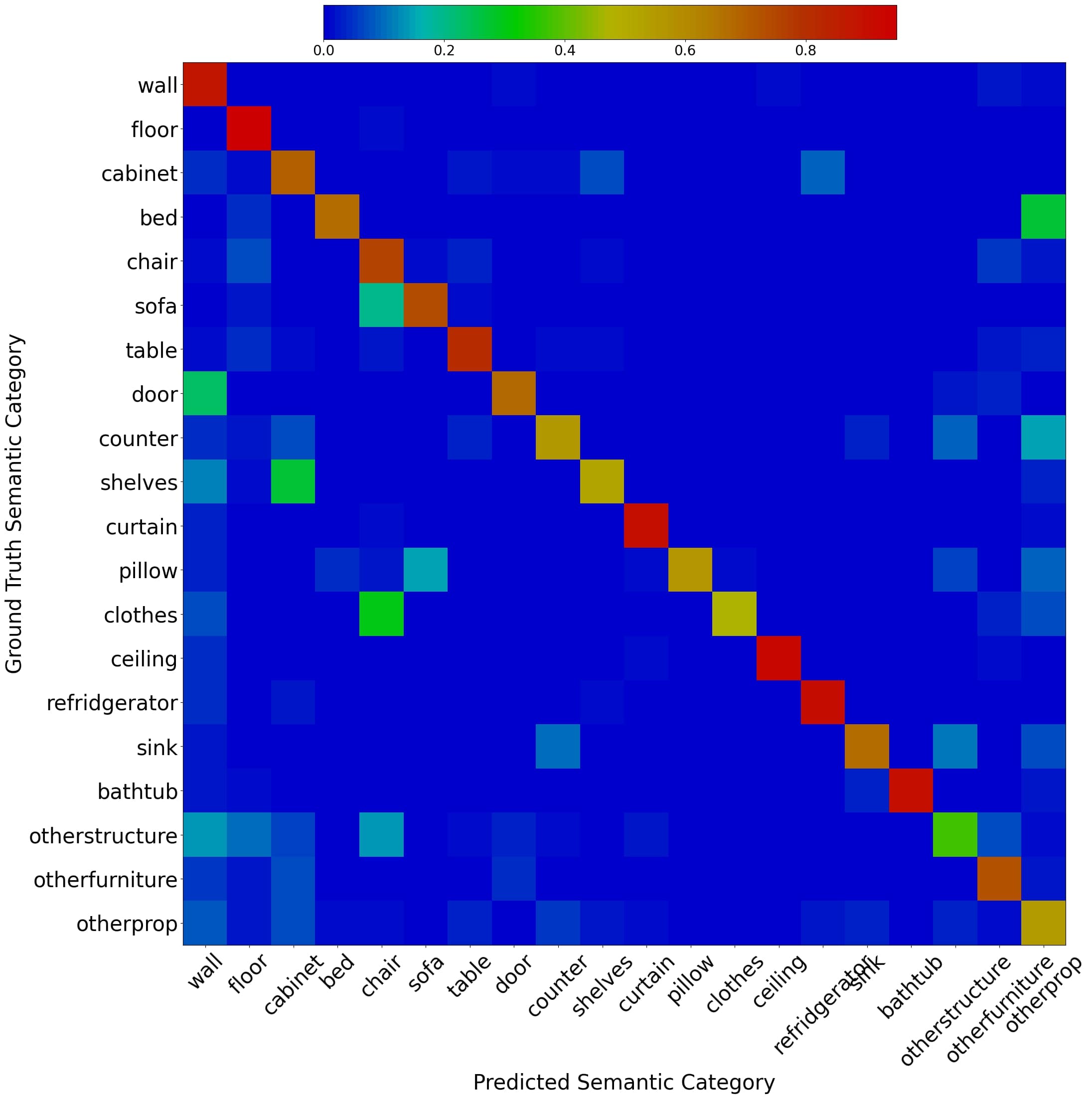} 
        \caption{$\mathcal{P} + \mathcal{A} + \mathcal{S}$}
        \label{fig:confugsion_matrices_sub3}
    \end{subfigure}
    \hfill 
    \begin{subfigure}[b]{0.45\textwidth}
        \centering
        \includegraphics[width=\textwidth]{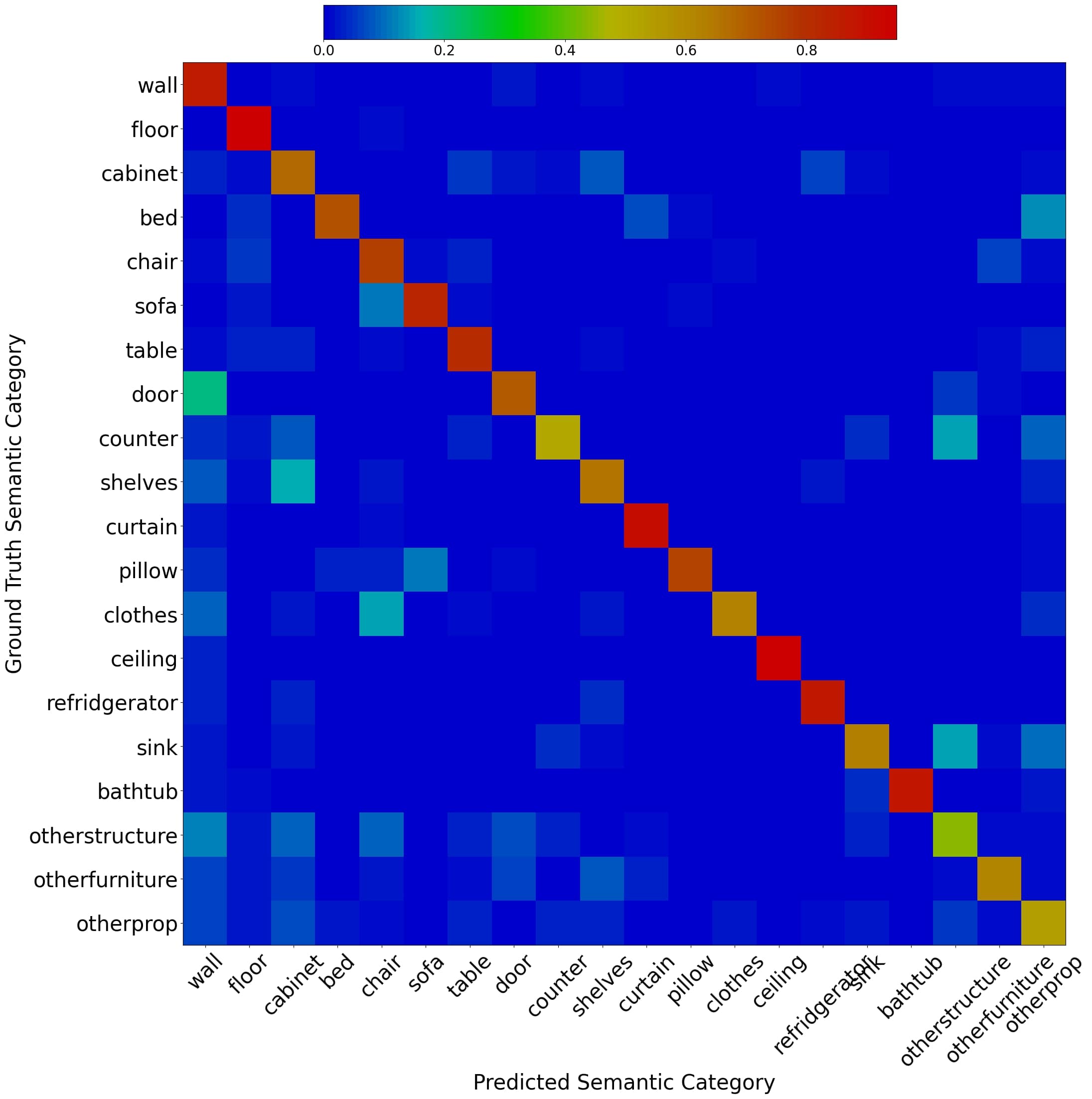} 
        \caption{$\mathcal{P} + \mathcal{A} + \mathcal{S} + \mathcal{R}$}
        \label{fig:confugsion_matrices_sub4}
    \end{subfigure}

    \begin{subfigure}[b]{0.45\textwidth}
        \centering
        \includegraphics[width=\textwidth]{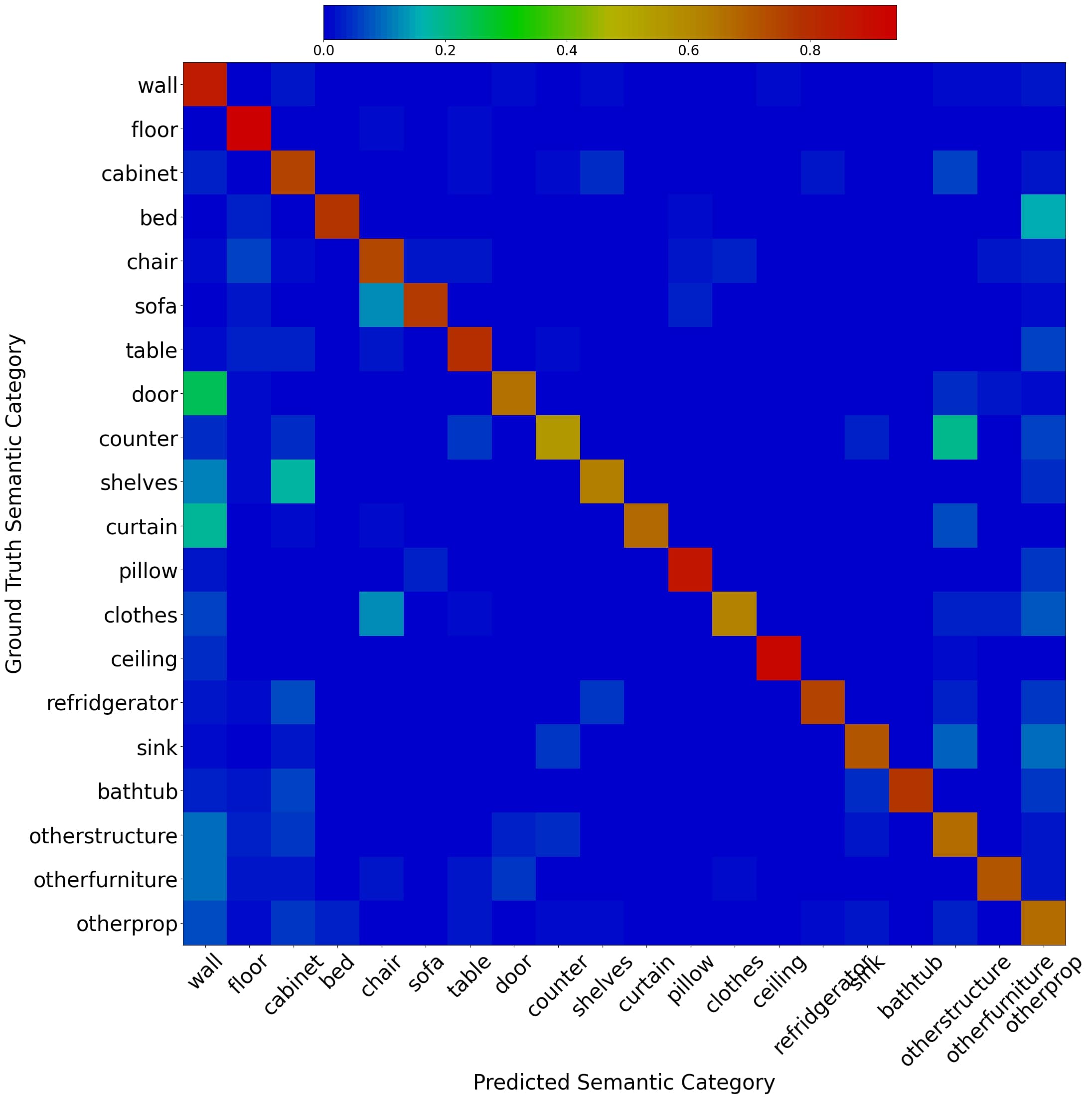} 
        \caption{$\mathcal{P} + \mathcal{I}$}
        \label{fig:confugsion_matrices_sub5}
    \end{subfigure}
    \hfill 
    \begin{subfigure}[b]{0.45\textwidth}
        \centering
        \includegraphics[width=\textwidth]{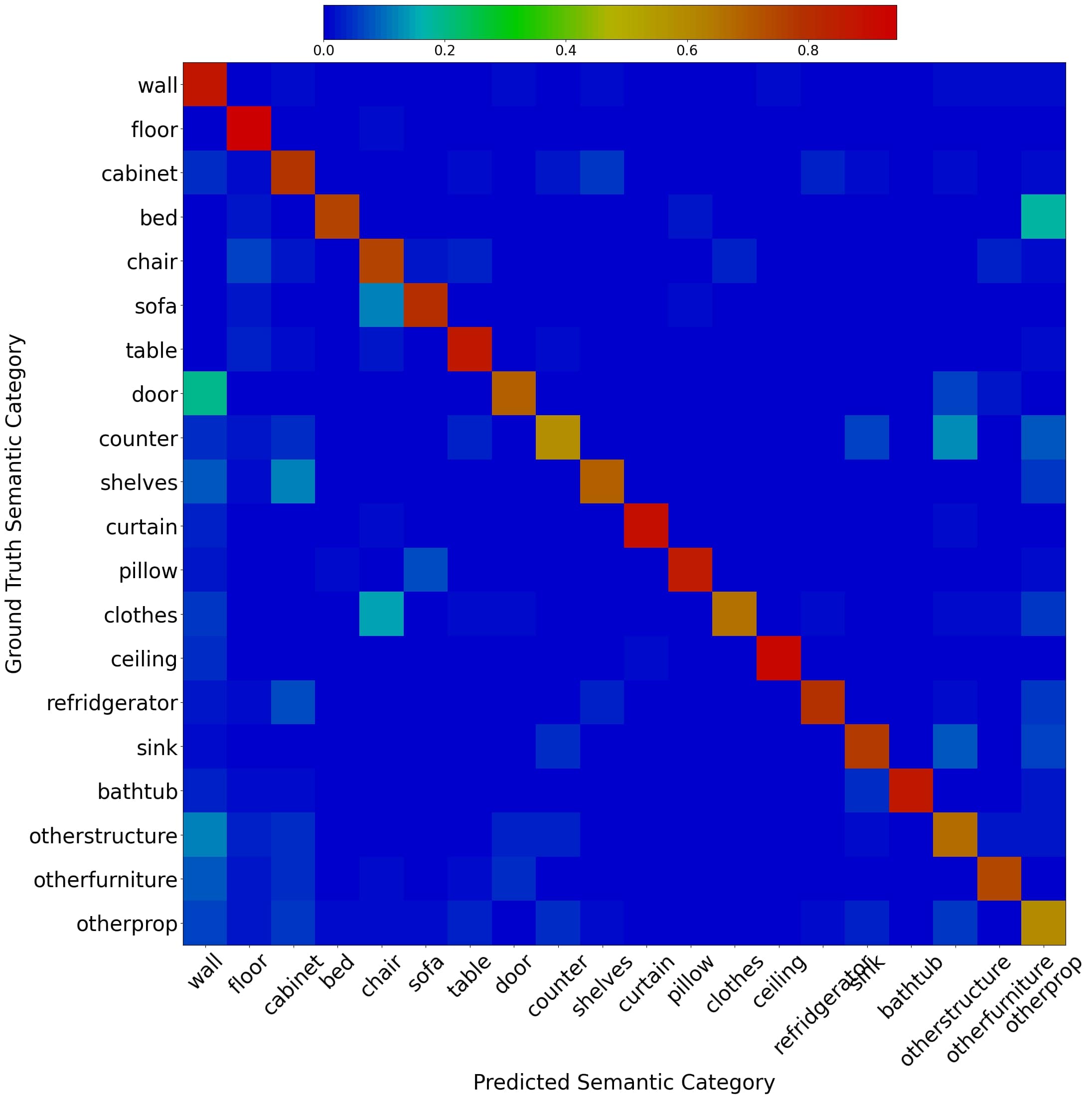} 
        \caption{$\mathcal{P} + \mathcal{I} + \mathcal{A} + \mathcal{S} + \mathcal{R}$}
        \label{fig:confugsion_matrices_sub6}
    \end{subfigure}
    \caption{Confusion Matrices of 6 modality combinations of \name{}. The y-axis represents the ground truth semantic category of the image patch $q$ of query image and the x-axis represents the semantic category of the object note $v$ in scene graph matched to $q$. Ideally, the confusion matrix should be the identity matrix.}
    \label{fig:confugsion_matrices}
\end{figure}

In \name{}, each patch of the query image $q \in \mathcal{Q}_I$ is assigned to an object node $v \in \mathcal{V}_I$ in the scene graph. We compute and visualize confusion matrices of semantic categories of ($q, v$) pairs, as illustrated in Fig.~\ref{fig:confugsion_matrices}. From the figure, we can see that as more modalities are integrated (from Fig.~\ref{fig:confugsion_matrices_sub1} to Fig.~\ref{fig:confugsion_matrices_sub6}), the confusion matrix is closer to the identity matrix, denoting that the patch-object matching becomes more accurate, which agrees with the trend of $Acc^t_q$ shown in Table~\ref{table: albation_statistics}. Fig.~\ref{fig:confugsion_matrices_sub6} shows that with all modalities integrated, there are still objects of certain categories with non-trivial probabilities of being mismatched: (i) image patches of \textit{counter} can be assigned to nodes of \textit{other structure}; (ii) patches of \textit{door} can be assigned to nodes of \textit{wall} and (iii) patches of clothes can be assigned to nodes of \textit{chair} due to the inaccurate instance segmentation when a cloth is hanging on the back of a chair, as shown in Fig.~\ref{fig:clothes_on_chair}.

\section{Implementation Details}
\label{sec: impl_details}

\textbf{Machine}. All the experiments of the room retrieval tasks during the inference phase are implemented on a machine with an Intel-12700K CPU, a Nvidia RTX3090 GPU and 64 GB RAM. 
For time measurement, the time $t_{e_q}$ of encoding the query image is measured by using the GPU and the time $t^{N}_\text{retr}$ of implementing room retrieval task is measured by using the CPU.
\par
\noindent\textbf{Models and Training}. 
We use $L=3, K_{view}=10$ for multi-level multi-view image embedding of objects as depicted in Section 3.1 in the main paper.
We use $\alpha = 0.5$ as the weight between static loss and temporal loss. 
The dimension $D^k$ for the embedding $e_v$ of each modality $k \in \{P, S, R, A\}$ is $100$ and the dimension for image modality is $256$. 
The dimension of unified embedding $D$ is $400$.
We train \name{} our model with a batch size of 16 using Adam~\cite{kingma2014adam} optimizer. Learning rate is 0.0011 with the step learning rate scheduler. 
\par
\noindent\textbf{Dataset}.
In 3RScan dataset~\cite{3rscan}, the query images are with resolution of $960\times540$ pixels and are resize to $224\times126$ pixels before feeding into the Dino~\cite{oquab2023dinov2} backbone, which then extract $16\times9$ patches features from the image. In ScanNet dataset~\cite{dai2017scannet}, images of $1296\times968$ are resized to $448\times338$ and $24\times32$ patch features are extracted. 
\clearpage
%
%
\bibliographystyle{splncs04}
\bibliography{egbib}

\end{document}